\documentclass[sigconf]{acmart}

\usepackage{graphicx}
\usepackage{lipsum} 
\usepackage{hyperref} 

\copyrightyear{2024}
\acmYear{2024}
\setcopyright{rightsretained}
\acmConference[KDD '24] {Proceedings of the 30th ACM SIGKDD Conference on Knowledge Discovery and Data Mining }{August 25--29, 2024}{Barcelona, Spain.}
\acmBooktitle{Proceedings of the 30th ACM SIGKDD Conference on Knowledge Discovery and Data Mining (KDD '24), August 25--29, 2024, Barcelona, Spain}
\acmDOI{10.1145/3637528.3671823}
\acmISBN{979-8-4007-0490-1/24/08}

\usepackage{lipsum}     
\usepackage{balance}    
\usepackage{booktabs}   
\usepackage{url}        
\usepackage{pifont}     
\usepackage{xcolor}      
\usepackage{amsmath,amssymb,amsfonts}
\usepackage{algorithm}
\usepackage{algpseudocode}
\usepackage{graphicx}
\usepackage{caption, subcaption}
\usepackage{textcomp}
\usepackage{color}
\usepackage{bm}
\usepackage{enumitem}
\usepackage{multirow}
\usepackage{colortbl,booktabs}
\usepackage{tabularx}
\definecolor{Gray}{gray}{0.9}

\usepackage{pbox}
\usepackage{makecell}
\usepackage{lipsum}
\usepackage{xcolor}
\usepackage{gensymb}
\usepackage{upgreek}
\usepackage{url}
\usepackage{multirow}
\usepackage{soul}
\usepackage{mathtools}
\usepackage{cuted}

\usepackage{xspace}

\makeatletter
\DeclareRobustCommand\onedot{\futurelet\@let@token\@onedot}
\def\@onedot{\ifx\@let@token.\else.\null\fi\xspace}

\def\eg{\emph{e.g}\onedot} 
\def\ie{\emph{i.e}\onedot} 
 
\def\etc{\emph{etc}\onedot}

\makeatother

\begin{document}


\title{Urban-Focused Multi-Task Offline Reinforcement Learning with Contrastive Data Sharing}


\author{Xinbo Zhao}
\affiliation{%
 \institution{Binghamton University}
 \city{Binghamton}
 \state{New York}
 \country{USA}}
 \email{xzhao9@binghamton.edu}

\author{Yingxue Zhang}
\affiliation{%
  \institution{Binghamton University}
  \city{Binghamton}
  \state{New York}
  \country{USA}}
  \email{yzhang42@binghamton.edu}

\author{Xin Zhang}
\affiliation{%
  \institution{San Diego State University}
  \city{San Diego}
  \state{California}
  \country{USA}}
  \email{xzhang19@sdsu.edu}

\author{Yu Yang}
\affiliation{%
  \institution{Lehigh University}
  \city{Bethlehem}
  \state{Pennsylvania}
  \country{USA}}
  \email{yuyang@lehigh.edu}

\author{Yiqun Xie}
\affiliation{%
  \institution{University of Maryland, College Park}
  \city{College Park}
  \state{Maryland}
  \country{USA}}
  \email{xie@umd.edu}

\author{Yanhua Li}
\affiliation{%
  \institution{Worcester Polytechnic Institute}
  \city{Worcester}
  \state{Massachusetts}
  \country{USA}}
  \email{yli15@wpi.edu}

\author{Jun Luo}
\affiliation{%
  \institution{Logistics and Supply Chain MultiTech R\&D Centre}
  \city{Hong Kong}
  \country{China}
}
  \email{jluo@lscm.hk}


\renewcommand{\shortauthors}{Xinbo Zhao et al.}

\begin{abstract}
Enhancing diverse human decision-making processes in an urban environment is a critical issue across various applications, including ride-sharing vehicle dispatching, public transportation management, and autonomous driving. Offline reinforcement learning (RL) is a promising approach to learn and optimize human urban strategies (or policies) from pre-collected human-generated spatial-temporal urban data. However, standard offline RL faces two significant challenges: (1) data scarcity and data heterogeneity, and (2) distributional shift.
In this paper, we introduce MODA — a \underline{M}ulti-Task \underline{O}ffline Reinforcement Learning with Contrastive \underline{D}ata Sh\underline{A}ring approach. MODA addresses the challenges of data scarcity and heterogeneity in a multi-task urban setting through Contrastive Data Sharing among tasks. This technique involves extracting latent representations of human behaviors by contrasting positive and negative data pairs. It then shares data presenting similar representations with the target task, facilitating data augmentation for each task.
Moreover, MODA develops a novel model-based multi-task offline RL algorithm. This algorithm constructs a robust Markov Decision Process (MDP) by integrating a dynamics model with a Generative Adversarial Network (GAN). Once the robust MDP is established, any online RL or planning algorithm can be applied.
Extensive experiments conducted in a real-world multi-task urban setting validate the effectiveness of MODA. The results demonstrate that MODA exhibits significant improvements compared to state-of-the-art baselines, showcasing its capability in advancing urban decision-making processes.
We also made our code available to the research community.

\end{abstract}

\ccsdesc[500]{Computing methodologies~Intelligent agents}
\ccsdesc[500]{Computing methodologies~Neural networks}
\keywords{Data Sharing, Offline Reinforcement Learning, Contrastive Learning}



\maketitle

\section{Introduction}\label{sec:intro}

\noindent{\bf Problem and Goal.}
In urban scenarios, individuals, referred to as human agents, engage in strategic planning to optimize their daily activities. These strategies devised to achieve specific objectives govern the human actions and decisions.
For example, taxi drivers strategize to maximize earnings and minimize travel time by selecting optimal pickup locations, adjusting working hours, and planning efficient routes.
Many urban strategies, such as those related to passenger seeking, public transit commuting and travel route selection, are individually crafted based on personal preferences. 
Consequently, they may not represent the most effective or efficient approaches and often require enhancement. 
In addition, these urban strategies are generally implicit to observers or even the human agents themselves, making their learning and improvement a complex task.
Therefore, in this paper, we aim to address this essential problem: \textit{How can we learn and enhance diverse human decisions in urban environments?}


\begin{figure}[t]
\includegraphics[width=0.45\textwidth]{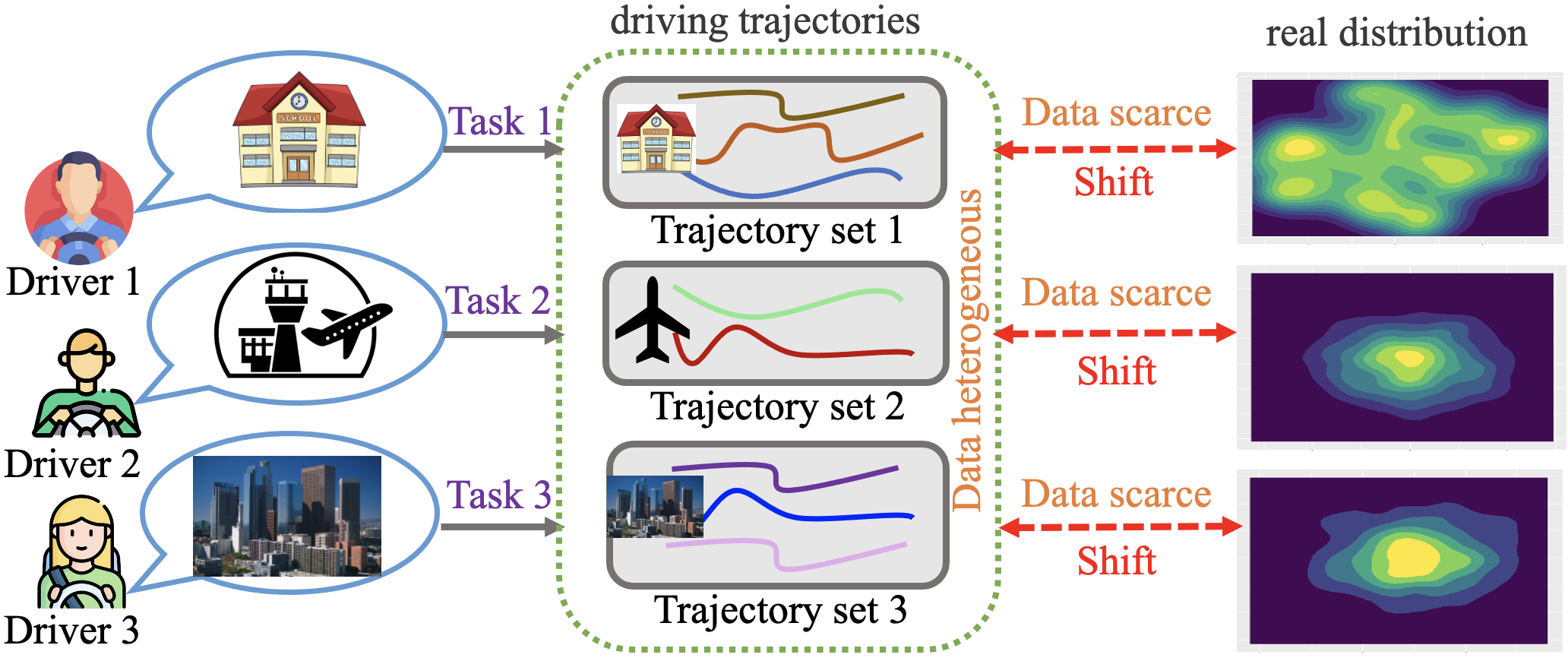}
\vspace{-0.35cm}
\caption{Example of data scarcity and heterogeneity, and distributional shift.
}
\label{fig:example}
\vspace{-15pt}
\end{figure}

\noindent{\bf Prior works and limitations.}
Different human strategies and decisions employed in various tasks lead to diverse human behaviors which are collected as human-generated spatial-temporal data (HSTD). Examples of HSTD include GPS trajectories from taxis and personal vehicles, route choices from bus and train passengers, \etc. 
Leveraging HSTD, data-driven techniques including imitation learning and offline reinforcement learning (RL) stand out as reliable tools for understanding human decision-making processes.
For instance, imitation learning and its variants ~\cite{gail,xgail,cgail,stmgail,traj_gail} aim at understanding human decision-making strategies and mimicking human behaviors across diverse urban scenarios using deep neural networks (DNNs). 
However, these approaches primarily focus on replicating human behaviors presented in the dataset, which is bounded by the true behavior policies that produce the dataset.
In this sense, the learned policies can hardly outperform true behavior policies, which renders them less suitable for enhancing human urban strategies with HSTD.
%
%
In contrast, offline RL~\cite{OfflineTutorial} leverages the inherent optimization mechanisms of RL to not only learn but also improve and optimize policies given HSTD. 
%
%
%
Many works have explored offline RL to solve urban problems. For example, some offline RL approaches are designed for traffic signal control~\cite{Kunjir2022,Dai2021,Li2023,anonymous2024motsc} and autonomous driving \cite{shah2022offline,Shi2021}.
However, these methods are generally limited to single-task scenarios within urban settings, as they can only address an individual task using a specific dataset, lacking the capability to learn from a composite dataset encompassing multiple tasks. 
This limitation restricts their applicability in more complex, multi-task urban environments.
Some multi-task offline RL approaches~\cite{CDS,UDS,Eysenbach2020} learn diverse skills for multiple tasks with various data-sharing techniques. However, these solutions are not transferrable to the urban domain as they rely on strong presumptions about the availability of explicit reward functions, which are inaccessible in most urban scenarios.

\begin{figure}[t]
\includegraphics[width=0.33\textwidth]{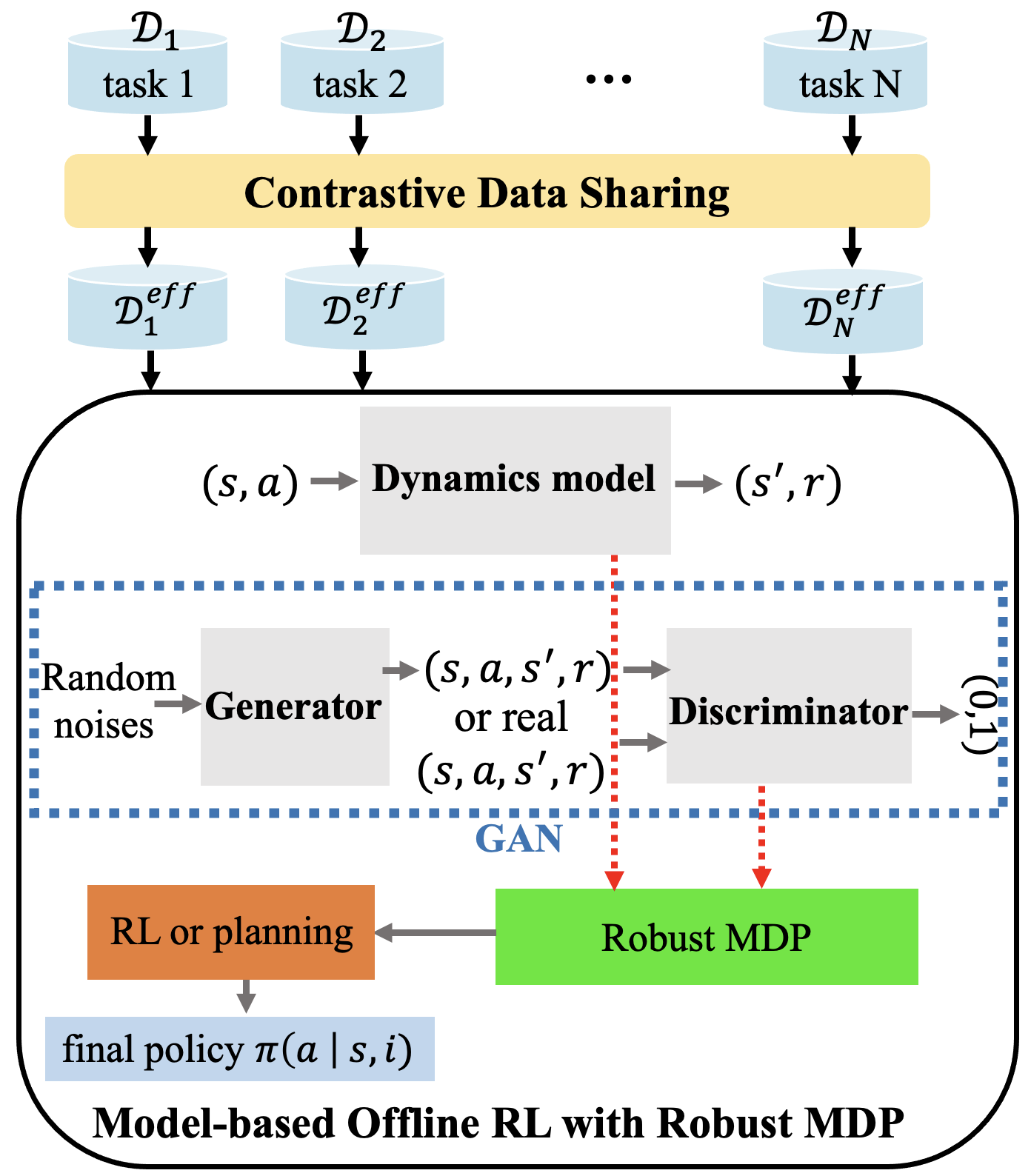}
\vspace{-0.35cm}
\caption{MODA overview.}
\label{fig:scenarios}
\vspace{-15pt}
\end{figure}

\noindent\textbf{The present work.} 
To address the above limitations and learn and enhance diverse human strategies,
we introduce MODA — a novel \underline{M}ulti-task \underline{O}ffline RL with Contrastive \underline{D}ata sh\underline{A}ring framework, which addresses the following two challenges:
\begin{itemize}[nosep, leftmargin=*]
    \item \textit{Data scarcity and data heterogeneity.} Learning from observations requires extensive amount of data from a task. However, in practice, it is difficult to collect a large amount of mobility data (\emph{i.e.}, HSTD) from each human agent. This leads to \textit{data scarcity}. 
    More commonly, the collected data originates from a variety of human agents, each employing a distinct urban strategy generating unique trajectories. This leads to \textit{data heterogeneity}. 
    Both data scarcity and heterogeneity problems complicate the process of learning and improving strategies for diverse human agents.

    \item \textit{Distributional shift.} Without online data collection during the learning process, offline RL often encounters failures attributed to substantial extrapolation errors when evaluating the $Q$-function on out-of-distribution actions~\cite{OfflineTutorial}. This issue worsens over the learning process as the learned policies increasingly deviate from the behavior policy. Moreover, the distributional shift problem is exacerbated in multi-task urban scenarios when learning urban strategies from diverse human agents, which leads to a huge divergence of the learned policies and poor performance.
\end{itemize}
As shown in Fig.~\ref{fig:scenarios}, MODA integrates a novel Contrastive Data Sharing method and an innovative model-based multi-task offline RL algorithm.
MODA first addresses the data scarcity and heterogeneity issues through Contrastive Data Sharing across all tasks. 
It then develops a novel model-based multi-task offline RL algorithm, which constructs a robust Markov Decision Process (MDP) by combining a dynamics model with a Generative Adversarial Network (GAN)~\cite{GAN}. 
This design can effectively mitigate the distributional shift challenge inherent in offline RL. 
\textit{The primary contributions of this paper can be summarized as follows:}

\begin{itemize}[nosep, leftmargin=*]
    \item We make the first attempt to optimize diverse human strategies by a novel multi-task offline RL framework named MODA.
    A key component of MODA is an innovative Contrastive Data Sharing strategy, which learns representations of human behaviors by contrasting positive and negative data pairs. 
    It then selectively shares data from other tasks that display patterns akin to those of the target task, ensuring a more effective sharing process to tackle the data scarcity and data heterogeneity problems.

    \item MODA also incorporates a novel model-based multi-task offline RL algorithm which can successfully address the distributional shift issue. It creates a robust MDP by learning a dynamics model and a GAN model from HSTD. 
    While the dynamics model generated transitions are not universally accurate --- given that the offline dataset after Contrastive Data Sharing still does not cover the entire state space --- the GAN's generator has a natural ability to generalize the offline dataset and learn the real MDP's data distribution. 
    This, in turn, improves the discriminator's ability to tell correct transitions from erroneous ones, making it more capable and generalizable. Once the learned dynamics model is integrated with the discriminator, which acts as a detector to distinguish reliable and out-of-distribution transitions, the accuracy of the dynamics model can be improved. Any online RL or planning algorithm can be applied once the robust MDP is constructed.

    \item Extensive experiments are conducted under a real-world multi-task urban setting with HSTD to validate the effectiveness of our MODA framework. The results demonstrate that MODA exhibits significant improvements compared to state-of-the-art baselines when learning different human urban strategies. {\em We also made our code available to the research community}\footnote{https://github.com/anony11sdf/MODA}.
    
\end{itemize}

\section{Overview}

In this section, we formally define the urban-focused multi-task offline RL problem. For ease of understanding and reference, all the notations used are listed in Table.~\ref{notation}.

\subsection{Preliminaries}

%
The standard RL framework treats sequential human decisions as a Markov Decision Process (MDP), often captured by HSTD in urban areas. 
HSTD encompasses the movements of diverse human agents, not just individual ones. In this paper, we treat the urban strategies of different agents as distinct tasks, 
and formalize 
the MDP as a 5-tuple $\mathcal{M}= (\mathcal{S}, \mathcal{A}, P, \gamma, \{r_i\}_{i=1}^N)$, where $\mathcal{S}$ and $\mathcal{A}$ denote the spaces of states and actions, respectively, $P(s^{\prime}| s, a)$ is the dynamics function, $\gamma\in [0, 1)$ is the discount factor, and $r_i$ represents the reward function for each task $i$ among a total of $N$ tasks, detailed further below. 

\par \noindent\textbf{Definition 1 (Grid cells).} 
A city is divided into $I\times J$ grid cells, with each cell having equal side-length in latitude and longitude. 
We denote the collection of grid cells in a city as $\mathcal{G}=\{g_{ij}\}$, where the indices $i$ and $j$ satisfy $1\le i\leq I$ and $1\le j\leq J$, respectively.


\par \noindent\textbf{Definition 2 (A state $\bm{s}$)}
represents a spatial-temporal location which
is defined as a tensor $\bm{s} \in \mathbb{R}^{n\times l\times l}$. 
The state $\bm{s}_{t}$ contains $n$ different feature maps $d\in \mathbb{R}^{l\times l}$ at time $t$. Since the human agents usually consider the status of the surrounding area when making urban decisions, each feature map characterizes a specific urban feature (\emph{e.g.}, traffic speed, travel demand, \emph{etc.}) of both the target grid cell and its neighboring $l\times l$ grid cells.


\par \noindent\textbf{Definition 3 (An action $a$)} 
is a decision made by a human agent at state $\bm{s}$. 
Following an action $a$, the human agent moves from a state $\bm{s}$ to the next state $\bm{s}^{\prime}$. 
We denote the set of actions as $\mathcal{A}=\{a\}$.

\par \noindent\textbf{Definition 4 (A dynamics $P$)} 
(or transition probability) is decided by the environment and is defined as $P: \mathcal{S} \times \mathcal{A} \times \mathcal{S}\mapsto [0,1]$. $P(s^\prime | s; a)$ characterizes the probability of transiting to state $\bm{s}^\prime$ at state $\bm{s}$ by following action $a$.

\par \noindent\textbf{Definition 5 (A reward function $r_i$)} 
is a mapping as $r_i: \mathcal{S} \times \mathcal{A} \mapsto \mathbb{R}$, which provides a numerical score based on a state $\bm{s}$ and an action $a$, and incentivizes a human agent $i$ to accomplish a task. Note that we treat distinct agents as separate tasks, using ``different agents'' and ``different tasks'' synonymously.   

\par \noindent\textbf{Definition 4 (A policy $\pi(a\mid s,i)$)} 
is a probability distribution defined as $\pi: \mathcal{S}\times \mathcal{A} \mapsto [0,1]$ indicating the probability of choosing an actions given the state $\bm{s}$. In this work, we follow \citet{CDS} and target on data sharing strategies to learn a conditional policy $\pi(a \mid \bm{s}, i)$ which indicates a policy for a target task $i$.

\par \noindent\textbf{Definition 5 (A trajectory $\tau$)} 
is a sequence of states and actions that a human agent traverses when completing a task in a geographic region, \ie, $\tau = (\bm{s}_0, a_0,\cdots, \bm{s}_T, a_T )$ of length $T$. 
The set of trajectories is denoted as $\mathcal{T}=\{\tau\}$. 
When generating these trajectories, the policy each human agent $i$ employs is named the behavior policy $\pi_\beta(a|\bm{s}, i)$.

\begin{table}[t]
\vspace{-0.4cm}
\caption{Notations}
\vspace{-0.4cm}
\begin{center}
\begin{tabular}{l|l}
\hline
\textbf{Notations}&\textbf{Descriptions}\\
\cline{1-2} 
$\mathcal{M}= (\mathcal{S}; \mathcal{A}; P; \gamma; \{r_i, i\}_{i=1}^N)$ & Multi-task Markov decision process. \\
\hline
$P(s^\prime | \bm{s}, a)$ & Dynamics function.  \\
\hline
$r_i(\bm{s},a)$ & Reward function for task $i$. \\
\hline
$\pi_i(a|\bm{s})$ or $\pi(a|\bm{s}, i)$ & Policy function for task $i$. \\
\hline
$\pi_\beta(a|\bm{s})$ & Behavior policy. \\
\hline
$\mathcal{D}_{i}^{\mathrm{eff}}$ & Effective dataset for task $i$. \\
\hline
$x_i$ & Sub-trajectory. \\
\hline
$f(x_i)$ & Latent representations of $x_i$.\\
\hline
\end{tabular}
\label{notation}
\end{center}
\vspace{-0.6cm}
\end{table}

\subsection{Urban-focused Multi-Task Offline RL Problem}
This paper aims to develop a new multi-task offline RL framework that can effectively optimize diverse human urban strategies while minimizing distributional shift and deal with the data scarcity and heterogeneity problems at the same time. The formal definition of the urban-focused multi-task offline RL problem is as follows:
\par \noindent\textbf{Problem Definition.} 
Given a set of human mobility trajectories $\mathcal{T}$ collected from $N$ different human agents, assume the data from each individual $i$ is generated by a specific behavior policy $\pi_\beta(a|\bm{s},i)$ to implement a certain task $i$,
we aim to learn the urban decision-making policy $\pi(a\mid \bm{s},i)$ for every agent so that they can produce higher expected cumulative rewards compared to the corresponding behavior policy $\pi_\beta(a\mid \bm{s},i)$.

%
%

\section{Methodologies}

In this section, we solve the urban-focused multi-task offline RL problem by introducing an innovative multi-task offline RL with contrastive data sharing approach, in short, MODA, as shown in Fig.~\ref{fig:scenarios}. 
MODA has two components to address the above challenges:\\
(1) {\em Contrastive Data Sharing.} 
To tackle the challenge of data scarcity and heterogeneity, we introduce a Contrastive Data Sharing mechanism. It enables data sharing across tasks by learning data representations and strategically sharing data from other tasks with the target task in a contrastive manner. The shared data is selected based on its similarity to the data of the target task, ensuring that they reflect comparable behaviors, preferences, and decision-making logic (See Section~\ref{sec:contrastive_sharing}). 
\\
(2) {\em Model-based multi-task offline RL with Robust MDP.} To tackle the challenge of distributional shift, we propose a novel model-based offline RL approach. 
We employ a GAN model~\cite{GAN} to enhance the reliability of the learned dynamics model, where the generator is trained to understand the data distribution of the actual environment and produce reliable transitions, while the discriminator distinguishes between reliable and out-of-distribution transitions.
A robust MDP is formed through the integration of the dynamics model and the discriminator. Any online RL or planning algorithm can be utilized once a robust MDP is established (See Section~\ref{sec:offlineRL}).
\\

\subsection{Contrastive Data Sharing}\label{sec:contrastive_sharing}
Contrastive Data Sharing is an important component in our MODA framework, which is specifically designed to address the data scarcity and heterogeneity challenge and thus enable effective data sharing among tasks. 
Standard multi-task offline RL usually requires large amounts of data for each task to ensure decent performance. However, HSTD is generally sourced from a variety of human agents or tasks, with each individual contributing a limited dataset. 
Thus, when attempting to learn different urban decision-making processes in a multi-task offline RL setting, we face data scarcity and heterogeneity problems.

\subsubsection{Limitations of state-of-the-art works.} Previous multi-task offline RL works \cite{Eysenbach2020,MT-Opt,CDS} have indicated the benefits of sharing data across multiple tasks to assist a target task $i$. However, these multi-task methods often presume direct access to the functional form of the reward $r_i$ for the target task $i$. 
In urban scenarios, reward functions and human preferences tend to be implicit, not only to external observers but also to the agents themselves, resulting in inaccessible reward functions. This lack of accessibility makes the prior multi-task offline RL techniques with data sharing ineffective for urban applications.

\begin{figure*}[t]
\includegraphics[width=0.9\textwidth]{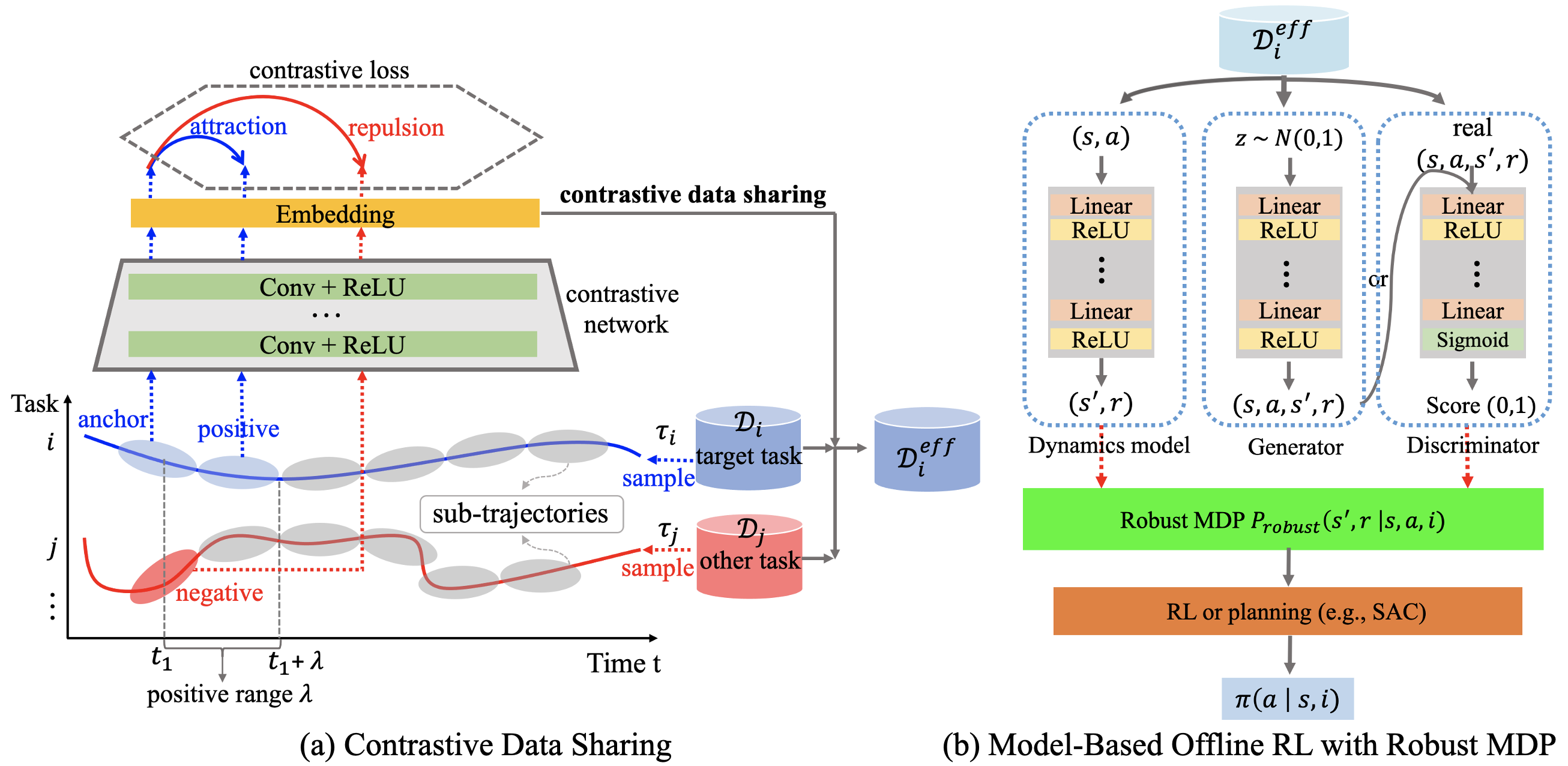}
\vspace{-0.35cm}
\caption{MODA structure. Figure 2(a) illustrates the detailed structure of Contrastive Data Sharing, which consists of a contrastive network working on positive and negative pairs of sub-trajectories from different tasks, a final embedding space is produced by following the contrastive loss. Figure 2(b) indicates the structure of the model-based multi-task offline RL with robust MDP, which is constructed by the combination of a dynamics model and the discriminator of a GAN.}
\label{fig:structure}
\vspace{-10pt}
\end{figure*}

\subsubsection{Contrastive Data Sharing Objective}
We introduce a novel data sharing method that employs contrastive learning to effectively augment the dataset for a target task by incorporating similar data from other tasks.
Consider a static multi-task dataset $\mathcal{D}=\bigcup_{i=1}^{N} \mathcal{D}_{i}$, where $N$ represents the number of tasks,
for a target task $i$, the data (typically a set of transitions $\{(\bm{s},a,\bm{s}^\prime, r)\}$) shared from task $j$ to task $i$ is denoted as $\mathcal{D}_{j \rightarrow i}$. Consequently, the augmented dataset for task $i$, termed as the effective dataset, is defined as: $\mathcal{D}_{i}^{\mathrm{eff}}:=\mathcal{D}_{i} \bigcup\left(\bigcup_{j \neq i} \mathcal{D}_{j \rightarrow i}\right)$.
The challenge then lies in identifying and sharing relevant data from other tasks to the target task $i$ effectively. 
Merely sharing all transitions indiscriminately among all tasks, a method we term \textit{Sharing All}, has been shown to yield suboptimal results in previous studies \cite{UDS,CDS}.
A more reasonable approach would be first 
learning the meaningful latent representations of trajectories collected from all tasks
and understanding the their similarities. Subsequently, only those trajectories from other tasks that exhibit behaviors similar to those of the target task $i$ are incorporated into the effective dataset $\mathcal{D}_{i}^{\mathrm{eff}}$.

Thus, to discern the similarities among the latent representations of all trajectories $\bigcup_{i=1}^{N}\{(\bm{s}_0,a_0,\cdots,\bm{s}_T,a_T)\}_i$, we design a Contrastive Data Sharing method. 
This method chooses a more nuanced approach for learning representations and data sharing by focusing on sub-trajectories rather than entire trajectories when sharing data from other tasks to the target task $i$. 
This is because the behaviors and decision-making patterns of human agents can significantly align in specific locations or time slots with certain segments of the target agent's trajectories. However, for a whole trajectory, behaviors and decision-making logic may vary considerably, leading to huge differences between complete trajectories.
Ignoring the commonalities within sub-trajectories would result in an inefficient data sharing process, as only a few trajectories from other tasks or agents would be considered relevant to the target task. 
By leveraging sub-trajectories, our method ensures a more effective data sharing process, which captures the nuanced similarities between agents' behaviors.

In Contrastive Data Sharing, a trajectory $\tau=(\bm{s}_0,a_0,\cdots,\bm{s}_T,a_T)$ is partitioned into multiple sub-trajectories 
$\tau =\{x_k\}$, where each $x_k$ represents a distinct sub-trajectory composed of $w$ consecutive transitions, \eg, $x_1=\{(\bm{s}_0,a_0,\bm{s}_1,r_1),\cdots,(\bm{s}_{w-1},a_{w-1},\bm{s}_w,r_w)\}$. 
The goal is to learn latent embeddings and discern similarities across all trajectories $\bigcup_{i=1}^{N}\{(\bm{s}_0,a_0,\cdots,\bm{s}_T,a_T)\}_i$ by examining the relationships between their sub-trajectories in a contrastive manner.
In this process, we construct positive samples as pairs of sub-trajectories from the same target task(/agent) exhibiting strong similarities, and negative samples as pairs of sub-trajectories from the target agent and other agents showing clear dissimilarities, as illustrated in Figure~\ref{fig:structure}(a). 
As decision-making processes in adjacent locations and time are often alike~\cite{stmgail}, sub-trajectories in close proximity typically exhibit similar behavior patterns. 
Given a trajectory from the target task, once we set a sub-trajectory as an anchor, we construct positive pairs composed of the anchor and another sub-trajectory within a specified positive range $\lambda$.
Conversely, sub-trajectories from different tasks at the same spatial-temporal states are considered negative samples due to behavioral differences.

The embedding of a sub-trajectory $x$ is represented by $f_\theta (x) \in \mathbb{R}^{d}$, where $f_\theta$ is the contrastive neural network parameterized by $\theta$.
To learn latent representations of sub-trajectories, Contrastive Data Sharing uses multi-view metric learning via a triplet loss~\cite{triplet_loss}, This loss ensures that a pair of sub-trajectories $x_{i}^{a}$ (anchor) and $x_{i}^{p}$ (positive) are closer to each other
in the latent space than any sub-trajectory $x_{j}^{n}$ (negative). Thus, we
aim to learn a contrastive network $f_\theta$ such that:
\begin{align}\label{eq:tcn}
\left\|f_\theta(x_{i}^{a})-f_\theta(x_{i}^{p})\right\|_{2}^{2}+\alpha<\left\|f_\theta(x_{i}^{a})-f_\theta(x_{j}^{n})\right\|_{2}^{2}, \\ 
\forall (x^{a}_i, x_{i}^{p}) \in \mathcal{D}_{i}, x_{j}^{n}\in \mathcal{D}\setminus\mathcal{D}_i, \nonumber
\end{align}
where $\alpha$ is a margin that is enforced between positive and negative pairs, and $\mathcal{D}_i$ is the dataset for target task $i$. The core idea is that two sub-trajectories (anchor and positive) coming from the target and showing a lot in common are pulled together, while a sub-trajectory from different tasks is pushed apart.

After learning the contastive network, the similarities between two sub-trajectories
can be evaluated by the Euclidean distance of their corresponding embeddings. The final data sharing strategy is as below:
\begin{align}\label{eq:sharing}
\mathcal{D}_{i}^{\mathrm{eff}}=\mathcal{D}_{i} \cup \{ x_j &|\text { if } \left\|\mu-f_\theta(x_{j})\right\|_{2}^{2} < 
\sigma^2,  \text { where } \mu = \frac{1}{m}\sum_{i=1}^{m}f_\theta(x_{i}^{a}),\nonumber\\   
 \sigma^2&=\frac{1}{m}\sum_{i=1}^{m}(f_\theta(x_i^{a})-\mu)^{2}, 
\ \forall x^{a}_i\in \mathcal{D}_{i}, x_j\in  \mathcal{D}\setminus\mathcal{D}_i\}.
\end{align}
Here, $m$ is the number of all possible anchors, $\sigma^2$ is the variance of the embeddings of all anchors from the target task $i$, $\mu$ is the average embedding of all anchors. Based on the sharing rule shown in 
Eq.~\eqref{eq:sharing}, if the similarities are smaller than 
the variance of the anchor, sub-trajectories $\{ x_j\}$ from other tasks can be added to the effective dataset $\mathcal{D}_{i}^{\mathrm{eff}}$ for target task $i$.
Note that once we get the final effective dataset $\mathcal{D}_{i}^{\mathrm{eff}}$, we break every sub-trajectory into transitions for future model-based multi-task offline RL.


\subsubsection{Structure \& Algorithm}
The contrastive network $f_\theta$ is composed of multiple convolutional layers activated by ReLU. The structure is illustrated in Figure~\ref{fig:structure}(a). The training process of the Contrastive Data Sharing is shown in Algorithm~\ref{alg:A}. 

\renewcommand{\algorithmicrequire}{ \textbf{Input:}}
\renewcommand{\algorithmicensure}{ \textbf{Output:}} 
\begin{algorithm}[h]
\caption{Contrastive Data Sharing Across Tasks}
\label{alg:A}
\begin{algorithmic}[1]
\Require{Datasets $\{\mathcal{D}_i\}$ and initialized $f_\theta$.}
\Ensure{Trained $f_\theta$ and the effective datasets $\{\mathcal{D}^{\mathrm{eff}}_i\}$ for all tasks.}
\For{$\text{task } i = 1,2,3,\cdots$}  
    \For{$\text{iteration } k = 1,2,3,\cdots$} 
    \State {Sample a batch of contrastive triples $\{(x_i^a,x_i^p,x_i^n)\}$.}
    \State {Get a batch of $\{(f(x_i^a),f(x_i^p),f(x_i^n))\}$ with $f_\theta$.}
    \State {Update $f_\theta$ with Eq.~\eqref{eq:tcn}.}
\EndFor
\State{Perform data sharing with Eq.~\eqref{eq:sharing} and get $\mathcal{D}^{\mathrm{eff}}_i$.}
\EndFor
\end{algorithmic}
\end{algorithm}

\subsection{Model-based Multi-task offline RL with Robust MDP}\label{sec:offlineRL}
After Contrastive Data Sharing, a novel model-based multi-task offline RL approach is designed to learn better policies compared to agents' behavior policies. The proposed model-based offline RL approach aims to construct a robust MDP by learning a dynamics model and a generative adversarial network (GAN). 
Once the dynamics model and the GAN's discriminator are integrated, a robust MDP can be established. 
Any state-of-the-art online RL or planning algorithm, such as Soft Actor-Critic (SAC)~\cite{SAC}, can be employed within this robust MDP to finally acquire the optimized policies for the target tasks.

\subsubsection{Robust MDP}
To successfully construct a robust MDP, we follow the steps including (i) learning a dynamics model, (ii) learning a GAN, and (iii) Robust MDP construction. 

\noindent\textbf{Learning a dynamic model.} 
In our proposed model-based offline RL algorithm, the first step involves using the offline dataset (\ie, the final effective dataset 
$\mathcal{D}_{i}^{\mathrm{eff}}$ for a target task $i$) to learn an approximate dynamics model with rewards $\hat{P}_\psi(\bm{s}^\prime, r | \bm{s}, a,i)$, where $\hat{P}_\psi$ is a deep neural network parameterized by $\psi$. Note we cannot use the whole dataset $\mathcal{D}$ to learn a $\hat{P}_\psi(\bm{s}^\prime, r | \bm{s}, a)$ applicable to all agents since different agents have different preferences and rewards, and correspond to different $\hat{P}_\psi(\bm{s}^\prime, r | \bm{s}, a,i)$\footnote{In practice, $\mathcal{D}$ is much larger than $\mathcal{D}_{i}^{\mathrm{eff}}$, if we learn a unified dynamics model $\hat{P}(\bm{s}^\prime | \bm{s}, a)$ using $\mathcal{D}$ and individual reward function $r_i(\bm{s},a)$ using $\mathcal{D}_{i}^{\mathrm{eff}}$,
$r_i(\bm{s},a)$ will not be compatible with $\hat{P}(\bm{s}^\prime | \bm{s}, a)$, and cannot produce reliable rewards for transitions not in $\mathcal{D}_{i}^{\mathrm{eff}}$, leading to a sub-optimal policy.}. This can be achieved through maximum likelihood estimation or other techniques from generative and dynamics modeling~\cite{Venkatraman2015, Samy2015, transformer}. 
However, urban environments are inherently complex and multifaceted. Even with the dataset obtained after Contrastive Data Sharing, it is impossible to cover the entire state space. 
Consequently, it is challenging to model every potential transition within the environment using the learned dynamics model. Thus, simply learning a dynamics model with rewards from the data $\mathcal{D}_{i}^{\mathrm{eff}}$ is far from enough to correctly reflect the real-world dynamics.

\noindent\textbf{Learning a GAN.} 
Given that the transitions generated by the learned dynamics model are not universally reliable, relying solely on this model could lead to a highly sub-optimal policy. This is primarily due to the model's potential to produce incorrect transitions, particularly in unexplored parts of the environment. 
Consequently, we utilize the discriminator of a GAN model to discern between reliable and unreliable transitions. Additionally, GAN's generator can generate more data based on the offline dataset.

By training a GAN using the dataset $\mathcal{D}_{i}^{\mathrm{eff}}$, the generator $G$ tries to learn the data distribution of $\mathcal{D}_{i}^{\mathrm{eff}}$. 
The goal of $G$ is to generate transitions that closely mimic the real transitions presented in the real MDP $\mathcal{M}$. The goal of discriminator $D$ is to differentiate between these generated transitions and the actual transitions from the dataset. 
When the generator tries to create transitions indistinguishable from real ones to effectively fooling the discriminator, the discriminator aims to enhance its ability to identify the generated transitions, thereby not being deceived. This adversarial process ensures continuous improvement of both components.

One significant advantage of incorporating a GAN in our model-based multi-task offline RL is the generator's inherent ability to generalize beyond the offline dataset. The generator will learn the data distribution of the real MDP and thus the generated transitions are more generalized and not only limited to  $\mathcal{D}_{i}^{\mathrm{eff}}$. This broader scope of generated transitions will benefit the discriminator as well, enhancing its capability in accurately distinguishing correct transitions from out-of-distribution ones, even for transitions that are not in $\mathcal{D}_{i}^{\mathrm{eff}}$. The objective of training the GAN model is listed as below:
\begin{align}
\min_{\phi}\max_{\chi} &\mathbb{E}_{(\bm{s},a,\bm{s}^\prime,r)\sim p_{data}}[\log D_\chi(\bm{s},a,\bm{s}^\prime,r)] \nonumber \\
& +\mathbb{E}_{z\sim p_z(z)}[\log(1-D_\chi(G_\phi(z)))],
\label{eq:GAN}
\end{align}
where $(\bm{s},a,\bm{s}^\prime,r)$ is a transition, $p_{data}$ is the data distribution of the real MDP, $z$ is a random noise sampled from a Gaussian. 
In this way, we get a discriminator working as the transition detector which can tell whether a transition is from the real data distribution or not. Both $G$ and $D$ are deep neural nets parameterized by $\phi$ and $\chi$.

\noindent\textbf{Robust MDP construction. }
Once we get the learned dynamics model and the discriminator, we combine them to construct a robust MDP. For every transition produced by the dynamics model, we use the discriminator to tell whether it is a reliable one. The final robust MDP is as below:
\begin{equation}
\hat{P}_{\text{robust}}\left(\bm{s}^{\prime},r | \bm{s}, a, i\right)=\left\{\begin{array}{ll}\delta\left(\bm{s}^{\prime}=\mathrm{Terminate}\right) & \text { if } D(\bm{s},a,\bm{s}^\prime,r)\ge\gamma, \\ 
\hat{P}\left(\bm{s}^{\prime},r | \bm{s}, a,i\right) & \text {if o.w, }   \end{array}\right.
\label{eq:MDP}
\end{equation}
where $\hat{P}_{\text{robust}}$ is the final robust MDP,
$\delta\left(\bm{s}^{\prime}=\mathrm{Terminate}\right)$ is the Dirac delta function, which forces the MDP to transit to the termination state, $\gamma$ is a threshold for the discriminator output (\ie, a score determining the reliability of a given transition),  a score higher than $\gamma$ indicates a transition is reliable, while a score below $\gamma$ suggests otherwise.

\subsubsection{Policy Optimization}
After the construction of the robust MDP, we treat each individual agent $i$ as a task under its robust MDP. Therefore, for each task $i$, we can apply any online RL or planning algorithms to optimize the generalized policy $\pi (a\mid \bm{s},i)$. 
In this paper, we directly apply Soft-Actor-Critic (SAC)~\cite{SAC} for policy optimization. 
The detailed algorithm for our model-based offline RL is shown in Algorithm~\ref{alg:B}. 

The structure of our model-based offline RL is shown in Figure~\ref{fig:structure}(b). 
The dynamics model $\hat{P}_\psi$ contains several fully-connected layers followed by ReLU. $\hat{P}_\psi$ uses a state $\bm{s}$ and an action $a$ as the input and tries to predict the next state $\bm{s}^\prime$ and reward $r$. The generator $G_\phi$ eats a random noise and generates a transition $(\bm{s},a,\bm{s}^\prime,r)$, the discriminator $D_\chi$ uses a real or generated transition $(\bm{s},a,\bm{s}^\prime,r)$ as input and outputs a score. Both $G_\phi$ and $D_\chi$ contains fully-connected layers followed by ReLU and Sigmoid, respectively.

\renewcommand{\algorithmicrequire}{ \textbf{Input:}}
\renewcommand{\algorithmicensure}{ \textbf{Output:}} 
\begin{algorithm}[h]
\caption{Model-Based Offline RL with Robust MDP}
\label{alg:B}
\begin{algorithmic}[1]
\Require{Effective dataset $\{\mathcal{D}^{\mathrm{eff}}_i\}$, initialized dynamics model $\hat{P}_\psi(\bm{s}^\prime, r| s,a,i)$, policy $\pi_\omega(a| \bm{s},i)$, generator $G_\phi$ and discriminator $D_\chi$.}
\Ensure{Robust MDP $\hat{P}_{\text{robust}}(\bm{s}^\prime, r| s,a,i)$ and well-learned  $\pi_\omega(a\mid \bm{s},i)$, $G_\phi$ and $D_\chi$.}
\For{$\text{task } i = 1,2,3,...$}
\For{$\text{iteration} = 1,2,3,...$}
    \State {Sample a batch of transitions $\{(\bm{s},a,\bm{s}^\prime,r)\}$ from $\mathcal{D}^{\mathrm{eff}}_i$.}
    \State {Update $\hat{P}_\psi$ with maximum likelihood estimation~\cite{MOPO}.}
\EndFor
\For{$\text{iteration} = 1,2,3,...$} 
    \State {Sample a batch of transitions $\{(\bm{s},a,\bm{s}^\prime,r)\}$ from $\mathcal{D}^{\mathrm{eff}}_i$.}
    \State {Update $G_\phi$ and $D_\chi$ with Eq.\ref{eq:GAN}.}
\EndFor
\State{Create $\hat{P}_{\text{robust}}$ using $\hat{P}_\psi$ and $D_\chi$ following Eq.~\eqref{eq:MDP}.}
\State{Get the policy $\pi_\omega$ by running SAC~\cite{SAC} in $\hat{P}_{\text{robust}}$.}
\EndFor
\end{algorithmic}
\end{algorithm}

\section{Experiments}
In our experiments, we aim to evaluate whether MODA can achieve satisfactory performance in the real-world multi-task urban setting. We will answer the following questions with extensive experiments: (1) Can our MODA learn good policies for different tasks compared to state-of-the-art baseline methods in the multi-task urban setting? (2) Can the Contrastive Data Sharing method in MODA effectively share data across tasks compared to other data sharing methods? (3) How do different features (including
the amount of shared data and other hyperparameters) affect the performance?

\begin{table*}[t]
\caption{Results on the taxi drivers’ passenger-seeking environment. All the
learned policies are evaluated using 20 rollouts in the simulated environments. The values are averaged over three random seeds.}
\vspace{-0.4cm}
\begin{center}
\setlength{\tabcolsep}{1.5mm}{
\begin{tabular}{c|c|c|c|c|c|c}
\hline
 & $\textbf{MODA}$ & CQL & BCQ  & BEAR & MOReL & MODA$-$ \\ \hline
Driver 1 (expert) & \textbf{135.54 $\pm$ 9.75} & 119.41 $\pm$ 9.10 & 110.94 $\pm$ 12.62 & 120.47 $\pm$ 10.63 & 109.28 $\pm$ 7.23 &   100.24 $\pm$ 5.34 \\ \hline
Driver 2 (medium) & \textbf{134.93 $\pm$ 9.95} & 117.40 $\pm$ 10.26 & 109.89 $\pm$ 16.72 & 116.08 $\pm$ 12.53 & 100.15 $\pm$ 6.38 & 96.75 $\pm$ 9.32  \\ \hline
Driver 3 (random)& \textbf{131.71 $\pm$ 6.23} & 122.70 $\pm$ 8.31 & 119.08 $\pm$ 9.25 & 121.82 $\pm$ 8.31 & 107.24 $\pm$ 9.66 &  100.96 $\pm$ 13.57 \\ \hline
\rowcolor{Gray} Average & \textbf{134.06 $\pm$ 8.81} & 117.33 $\pm$ 11.99 & 115.81 $\pm$ 10.80 & 119.46 $\pm$ 10.63 &  105.56 $\pm$ 7.88  & 99.32 $\pm$ 9.99 \\ \hline
\end{tabular}
\label{task1}
}
\end{center}
\end{table*}

\subsection{Dataset and Experiment Descriptions}
\textbf{Data description.} We evaluate our MODA using the taxi trajectory dataset representing diverse taxi drivers’ passenger-seeking strategies. Different drivers employ different strategies for seeking passengers which can be viewed as different tasks.
The passenger-seeking trajectories are collected from 17,877 taxis in Shenzhen, China from July 1 to Sep 31, 2016.
Each passenger-seeking trajectory is formed by multiple consecutive GPS records of a certain taxi. A GPS record includes five attributes including the taxi plate ID, longitude, latitude, time stamp and passenger indicator.

In this dataset, the whole Shenzhen City is first divided into $40\times50$ equal-sized grid cells with a side-length $l_1 = 0.0084^\circ$ in latitude and $l_2 = 0.0126^\circ$ in longitude. And the time of a day is divided into five-minute time slots. A spatial-temporal state is defined as a multi-dimensional tensor which is composed of different feature maps of its neighboring 5 × 5 grid cells in a specific time slot, here the features includes traffic volume, travel demand, traffic speed, waiting time, and the distance to the selected Points of Interests (PoIs).
When a taxi is in a specific state, the taxi driver has 10 actions to choose from, including going to 8 neighboring grid cells, staying at the current grid cell, and terminating the trip.

\noindent\textbf{Passenger-seeking simulation:} We created a simulated passenger-seeking environment based on the above taxi trajectory data for the comparison of approaches requiring environmental interactions. We have opened source this simulation environment.
More details about data and the simulated environment can be found in Appendix.

\subsection{Baselines}
To evaluate our proposed model-based offline RL with robust MDP, we evaluate multiple state-of-the-art offline RL models as baselines including model-free algorithms \textbf{CQL}~\cite{CQL}, \textbf{BCQ}~\cite{Fujimoto2018}, \textbf{BEAR}~\cite{Kumar2019}, and a model-based algorithm \textbf{MOReL}~\cite{MOReL}. All algorithms are running on the same effective datasets after applying data sharing for target tasks. Besides, a baseline called \textbf{MODA}$-$ is also compared which removes the GAN from our MODA.
Furthermore, to validate whether our Contrastive Data Sharing method is effective, we compare it with other data sharing strategies without requiring reward functions including \textbf{No Sharing}~\cite{CDS}, \textbf{Sharing All}~\cite{CDS}, and \textbf{UDS}~
\cite{UDS}. No Sharing doesn't perform any data sharing, Sharing All naively shares all data across tasks, UDS relabels data from other tasks with zero reward and then share with the target.

\begin{figure}[t]
\includegraphics[width=0.45\textwidth]{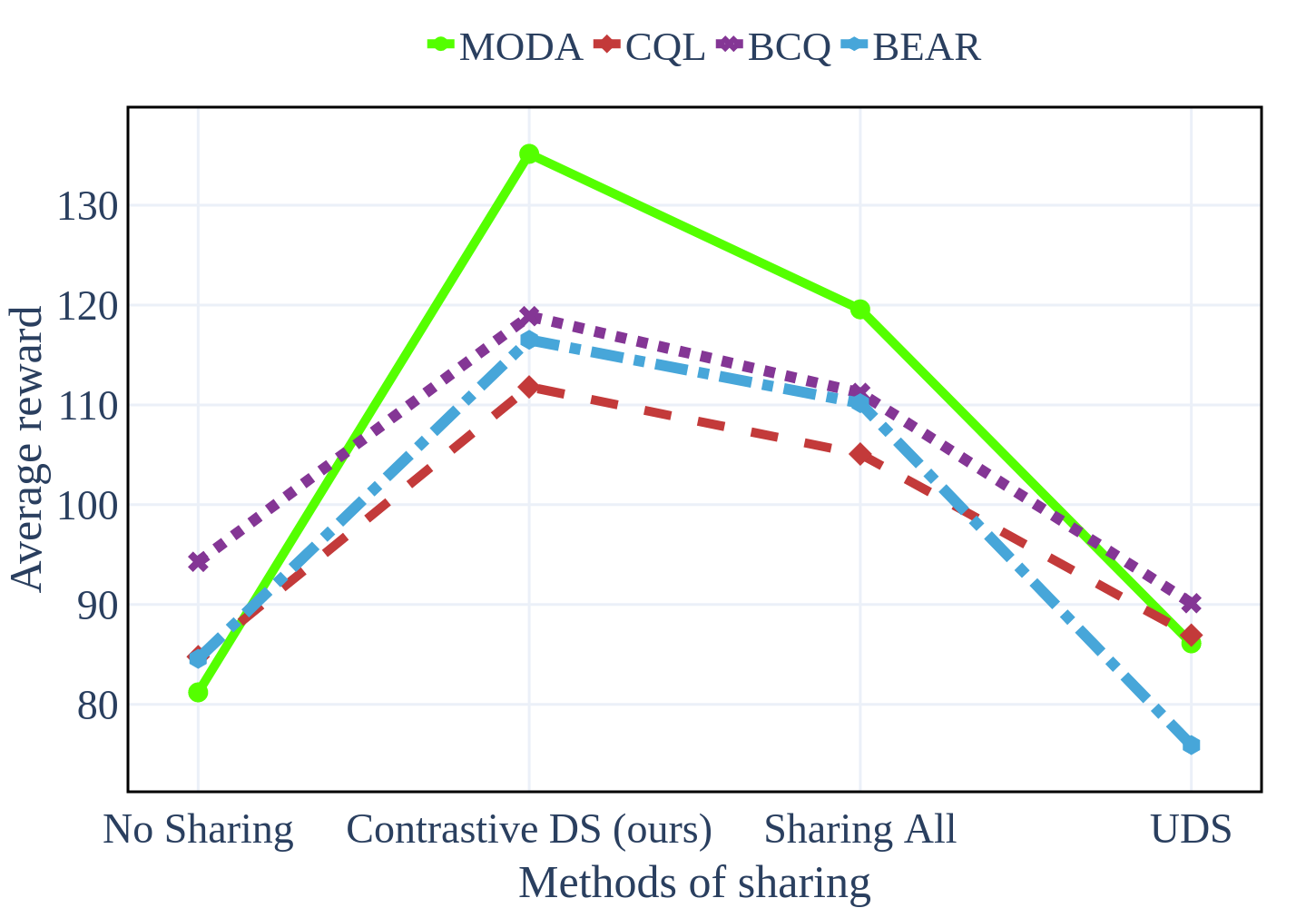}
\vspace{-12pt}
\caption{Performance of Offline RL algorithms with different data sharing approaches. All results are averaged over 20 rollouts, utilizing three
random seeds.
}
\label{task2}
\vspace{-20pt}
\end{figure}

\subsection{Experimental Settings}

\noindent\textbf{Contrastive Data Sharing}. We train the Contrastive learning models by randomly sampling triples containing Anchor, positive, and negative from the dataset. We use 4 convolutional layers with kernel size 3. The batch size is set to 32. Adam optimizer is applied.  
All models are trained for 2000 epochs.

\noindent\textbf{Model-based Multi-task offline RL with Robust MDP}. We use transitions from effective dataset $\mathcal{D}_{i}^{\mathrm{eff}}$ to train GAN and the dynamics model. Then we use the discriminator in GAN and dynamics model to build a Robust MDP. 
More details about these two models and Robust MDP can be found in the Appendix.

\begin{figure}[t]
\includegraphics[width=0.45\textwidth]{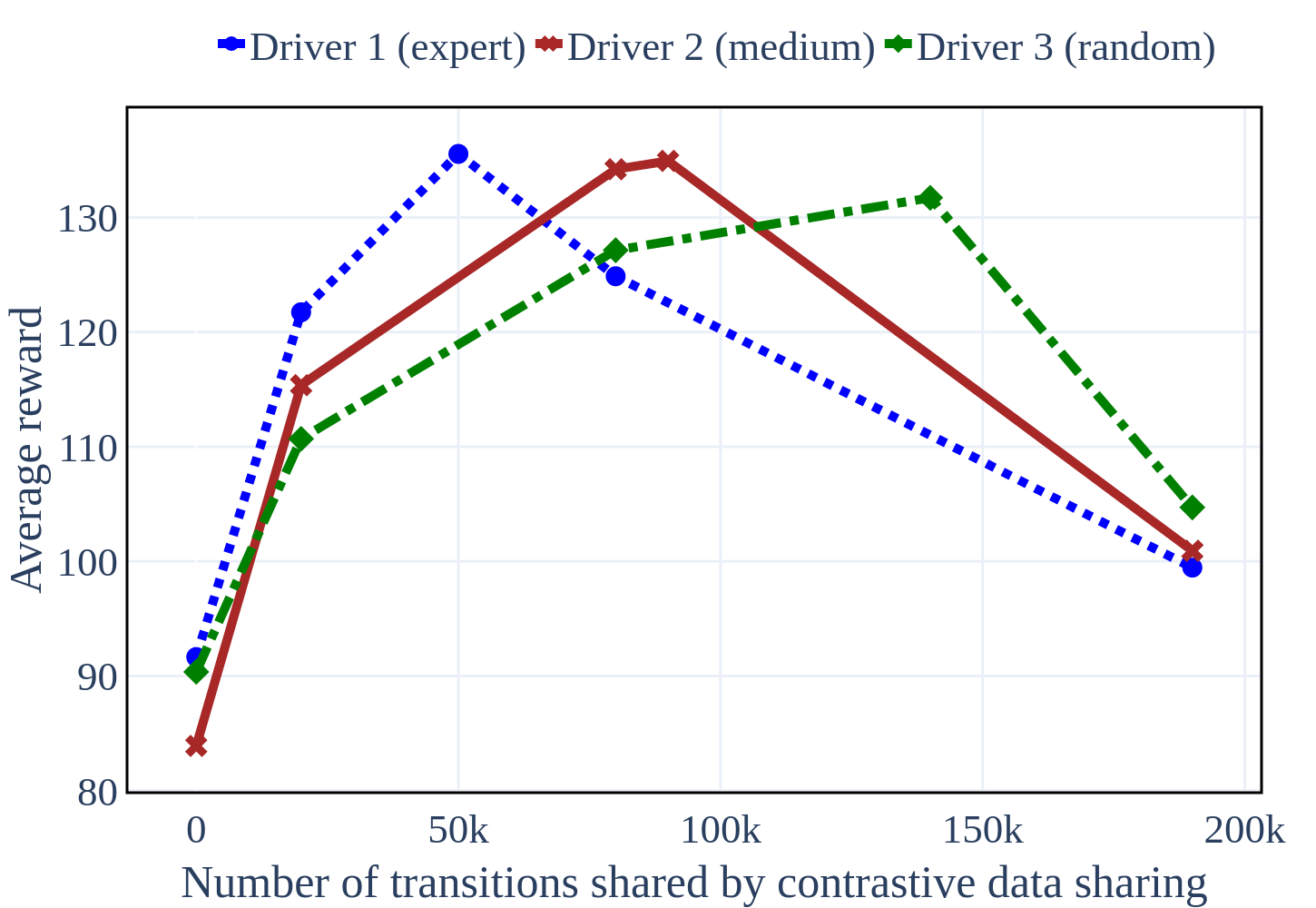}
\vspace{-0.35cm}
\caption{Impact of different number of transitions shared by Contrastive Data Sharing.
MODA exhibits varying performance across drivers of different expertise levels when a distinct number of transitions are shared with the target driver. All results are averaged over 20 rollouts, utilizing three random seeds.
}
\label{task3}
\vspace{-15pt}
\end{figure}

\begin{figure*}[h]
\includegraphics[width=1\textwidth]{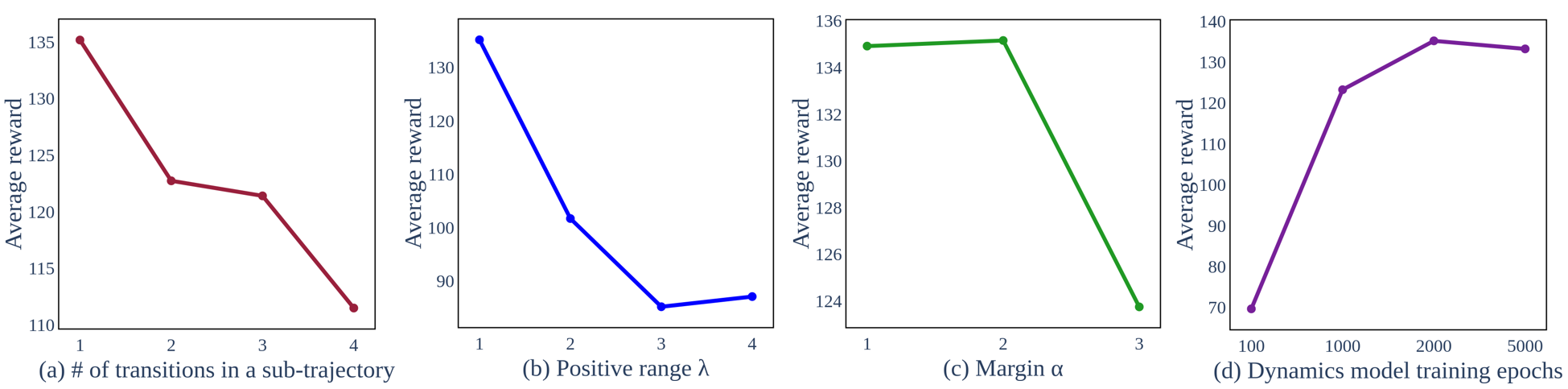}
\vspace{-20pt}
\caption{Impact of hyper-parameters on MODA.
}
\label{hyper}
\vspace{-15pt}
\end{figure*}

\subsection{Empirical Results}
\noindent\textbf{Results of Question (1)}.
To evaluate whether our proposed model-based offline RL with robust MDP in MODA can produce good policies for different tasks, we compare MODA with baselines including CQL, BCQ, BEAR, MOReL and MODA$-$ for 3 different drivers. All baselines are built upon the same effective dataset for a specific driver after contrastive data sharing. As shown in Table.~\ref{task1}, we pick 3 drivers from our dataset including an expert driver, a medium-skilled driver and a random driver, whose behavior policies are optimal, medium and random policies, respectively.  Each value in Table.~\ref{task1}
is the average expected cumulative reward for 20 rollouts in the simulated environment over three random seeds. We also present the average results of all drivers for different models. 
The findings, as detailed in Table~\ref{task1}, demonstrate that MODA significantly surpasses the baselines, substantiating the effectiveness of our robust MDP construction and our approach's ability to address distributional shifts.
Notably, MODA$-$, which excludes the GAN component from MODA, exhibits inferior performance, underscoring the critical role of GAN in achieving a robust MDP within MODA. 
Furthermore, MODA can successfully enhance policy performance across drivers of varying skill levels, with even medium and random drivers whose behavior policies are highly suboptimal achieving outcomes comparable to those of the expert driver. This underscores MODA's capacity to markedly improve policy effectiveness across diverse tasks.

\noindent\textbf{Results of Question (2)}.
To assess the efficacy of the Contrastive Data Sharing (DS) method in MODA, we conducted a comparative analysis against alternative data sharing strategies, including No Sharing, Sharing All, and UDS. 
As shown in Figure~\ref{task2}, for a certain driver, we run different offline RL models including MODA, CQL, BCQ and BEAR on the effective datasets constructed by Contrastive DS, No Sharing, Sharing All and UDS. We find 
No Sharing leads to poor performance mainly due to the lack of data for the target driver. Naively Sharing All will introduce a lot of bias to the target driver and result in low rewards. UDS relabels the rewards before sharing data which degrades the RL performance. In contrast,
all offline RL models achieve the highest rewards on the effective dataset built by Contrastive Data Sharing, which indicates our Contrastive Data Sharing can successfully share similar data from other tasks with the target task by learning effective data representations,
ensuring that the shared data reflect comparable behaviors,
preferences, and decision-making logic.

\noindent\textbf{Results of Question (3)}. We also evaluate how different amounts of data shared with the target task by Contrastive Data Sharing affect the MODA performance. As shown in Figure~\ref{task3}, we find MODA exhibits varying performance across drivers of different expertise levels when a distinct number of transitions are shared with the target driver. For example, the expert driver needs the least number of shared data and achieves the best performance, and the random driver needs the most number of shared data to achieve a better performance. This results align with the real-world cases where it's much easier for an expert driver to get high rewards if the shared data is effective, and for random drivers, they usually need more effective data to improve their policies. 

Besides, we also evaluate how different hyperparameters including the number of transitions in a sub-trajectory, positive range $\lambda$, the margin $\alpha$ and the training epochs affect our MODA performance. As shown in Figure~\ref{hyper}(a), we find the MODA performance decreases if a sub-trajectory contains more transitions, since a long sub-trtajectory will contain different spatial-temporal states and actions and thus introduce more bias to the model. As shown in Figure~\ref{hyper}(b), we find large positive range $\lambda$ will degrade the performance since there is no guarantee that positive pairs will exhibit similar behaviors. As shown in Figure~\ref{hyper}(c), if margin is too large, we cannot guarantee an effective contrastive loss. As shown in Figure~\ref{hyper}(d), more training epochs will improve the quality of the dynamics model.

\section{Related Work}
\noindent\textbf{Offline RL.} 
Offline RL~\cite{OfflineTutorial,batchRL,batchRL2} is a data-driven RL framework that learns better policies from previously collected datasets without further environment interaction. It shows promising performance across various domains, including robotic control~\cite{Kalashnikov2018,Rafailov2020}, NLP~\cite{Jaques2019}, and healthcare~\cite{Wang2018}. However, offline RL faces the challenge of distributional shift, where the learned policy diverges from the behavior policy, leading to erroneous value backups~\cite{Fujimoto2018,OfflineTutorial}. 
To address this problem, some offline RL algorithms introduce constraints on the policy \cite{Kumar2019,Fujimoto2018,Wu2019,Jaques2019,Peng2019,Siegel2020} or on the learned Q-function \cite{CQL,Kostrikov2021}, to ensure the learned policy does not stray excessively from the behavior policy. Additionally, model-based offline RL strategies have been explored to mitigate distributional shift, such as constructing a conservative MDP \cite{MOReL} or implementing reward penalties \cite{MOPO}. However, all these works are not applicable in our multi-task urban settings since they did not consider the practical data issues (\emph{e.g.}, data scarcity and heterogeneity) and the spatial-temporal nature of urban decisions.

\noindent\textbf{Multi-Task RL.} 
Multi-task RL learns multiple tasks simultaneously to improve generalization and learning efficiency in RL settings. 
The prior work\cite{tessler2017deep} utilize the transfer learning concept,  which shares a "distilled" policy across tasks to encourage positive transfer while maintaining task-specific policies. 
More recent approaches have focused on leveraging meta-learning for multi-task RL. Finn et al. \cite{MAML} introduced Model-Agnostic Meta-Learning (MAML), a versatile framework applicable to multi-task RL, demonstrating substantial improvements in learning efficiency. Similarly, Hessel et al. \cite{hessel2019multi} proposed the PopArt normalization technique to stabilize value estimates across varying reward scales in multi-task settings.  Besides, the works~\cite{MultiOffline1,MultiOffline2,CDS,UDS} related to multi-task offline RL mainly focus on goal-conditioned RL. In contrast, in this paper, we focus on the multi-task urban setting.

\noindent\textbf{RL with data sharing.} 
Sharing data across tasks has been demonstrated as highly beneficial in both multi-task RL~\cite{Eysenbach2020,CDS,MT-Opt} and meta-RL~\cite{OfflineMeta1,OfflineMeta2,OfflineMeta3}. Prior works have explored various data-sharing approaches, such as utilizing learned Q-values~\cite{Eysenbach2020,CDS}, domain knowledge~\cite{MT-Opt}, or distances to goals in goal-conditioned RL settings~\cite{Liu2019,Lin2019,Sun2019,Chebotar2021}. Another approach UDS~\cite{UDS} treats all unlabeled data as having zero reward to facilitate data sharing. Moreover, most of these works rely on access to the reward function of the target task and exhibit limitations in urban scenarios. In contrast, our work does not require any reward function and tries to share data based on latent similarities.

\section{Conclusion}
In this paper, we aim to enhance diverse human decision-making processes in an urban environment using offline RL which can learn policies from a static dataset pre-collected by a certain behavior policy. However, standard offline RL faces two significant challenges: (i) data scarcity and heterogeneity, and (ii) distributional shift.
To address both challenges in a multi-task offline RL setting where learning the policy for each human agent can be viewed as a task, we introduce MODA — a Multi-Task Offline Reinforcement Learning with Contrastive Data Sharing approach. MODA includes a new contrastive data sharing method which can extract latent representations of human behaviors by contrasting positive and negative data pairs. 
Moreover, MODA develops a novel model-based offline RL algorithm. This algorithm constructs a robust MDP by integrating a dynamics model with a Generative Adversarial Network (GAN). Once the robust MDP is established, any online RL or planning algorithm can be applied.
Extensive experiments conducted in a real-world multi-task urban setting has validated the effectiveness of MODA.



\begin{acks}
This work was supported in part by NSF grants IIS-1942680 (CAREER), CNS-1952085 and DGE-2021871. Jun Luo is supported by The Innovation and Technology Fund (Ref. ITP/069/23LP).\\
Disclaimer:
Any opinions, findings, conclusions or recommendations expressed in this material/event (or by members of the project team) do not reflect the views of the Government of the Hong Kong Special Administrative Region, the Innovation and Technology Commission or the Innovation and Technology Fund Research Projects Assessment Panel.
\end{acks}

\balance
\bibliographystyle{ACM-Reference-Format}
\bibliography{reference}


\begin{thebibliography}{49}


\ifx \showCODEN    \undefined \def \showCODEN     #1{\unskip}     \fi
\ifx \showDOI      \undefined \def \showDOI       #1{#1}\fi
\ifx \showISBNx    \undefined \def \showISBNx     #1{\unskip}     \fi
\ifx \showISBNxiii \undefined \def \showISBNxiii  #1{\unskip}     \fi
\ifx \showISSN     \undefined \def \showISSN      #1{\unskip}     \fi
\ifx \showLCCN     \undefined \def \showLCCN      #1{\unskip}     \fi
\ifx \shownote     \undefined \def \shownote      #1{#1}          \fi
\ifx \showarticletitle \undefined \def \showarticletitle #1{#1}   \fi
\ifx \showURL      \undefined \def \showURL       {\relax}        \fi
\providecommand\bibfield[2]{#2}
\providecommand\bibinfo[2]{#2}
\providecommand\natexlab[1]{#1}
\providecommand\showeprint[2][]{arXiv:#2}

\bibitem[ano(2024)]%
        {anonymous2024motsc}
 \bibinfo{year}{2024}\natexlab{}.
\newblock \bibinfo{title}{{MOTSC}: Model-based Offline Traffic Signal Control}.
\newblock
\newblock
\urldef\tempurl%
\url{https://openreview.net/forum?id=K6BXvqWWmq}
\showURL{%
\tempurl}


\bibitem[Andrychowicz et~al\mbox{.}(2017)]%
        {MultiOffline1}
\bibfield{author}{\bibinfo{person}{Marcin Andrychowicz}, \bibinfo{person}{Filip Wolski}, \bibinfo{person}{Alex Ray}, \bibinfo{person}{Jonas Schneider}, \bibinfo{person}{Rachel Fong}, \bibinfo{person}{Peter Welinder}, \bibinfo{person}{Bob McGrew}, \bibinfo{person}{Josh Tobin}, \bibinfo{person}{Pieter Abbeel}, {and} \bibinfo{person}{Wojciech Zaremba}.} \bibinfo{year}{2017}\natexlab{}.
\newblock \showarticletitle{Hindsight Experience Replay}.
\newblock \bibinfo{journal}{\emph{CoRR}}  \bibinfo{volume}{abs/1707.01495} (\bibinfo{year}{2017}).
\newblock
\showeprint[arXiv]{1707.01495}
\urldef\tempurl%
\url{http://arxiv.org/abs/1707.01495}
\showURL{%
\tempurl}


\bibitem[Bengio et~al\mbox{.}(2015)]%
        {Samy2015}
\bibfield{author}{\bibinfo{person}{Samy Bengio}, \bibinfo{person}{Oriol Vinyals}, \bibinfo{person}{Navdeep Jaitly}, {and} \bibinfo{person}{Noam Shazeer}.} \bibinfo{year}{2015}\natexlab{}.
\newblock \bibinfo{title}{Scheduled Sampling for Sequence Prediction with Recurrent Neural Networks}.
\newblock
\newblock
\showeprint[arxiv]{1506.03099}~[cs.LG]


\bibitem[Chebotar et~al\mbox{.}(2021)]%
        {Chebotar2021}
\bibfield{author}{\bibinfo{person}{Yevgen Chebotar}, \bibinfo{person}{Karol Hausman}, \bibinfo{person}{Yao Lu}, \bibinfo{person}{Ted Xiao}, \bibinfo{person}{Dmitry Kalashnikov}, \bibinfo{person}{Jake Varley}, \bibinfo{person}{Alex Irpan}, \bibinfo{person}{Benjamin Eysenbach}, \bibinfo{person}{Ryan Julian}, \bibinfo{person}{Chelsea Finn}, {and} \bibinfo{person}{Sergey Levine}.} \bibinfo{year}{2021}\natexlab{}.
\newblock \showarticletitle{Actionable Models: Unsupervised Offline Reinforcement Learning of Robotic Skills}.
\newblock \bibinfo{journal}{\emph{CoRR}}  \bibinfo{volume}{abs/2104.07749} (\bibinfo{year}{2021}).
\newblock
\showeprint[arXiv]{2104.07749}
\urldef\tempurl%
\url{https://arxiv.org/abs/2104.07749}
\showURL{%
\tempurl}


\bibitem[Dai et~al\mbox{.}(2021)]%
        {Dai2021}
\bibfield{author}{\bibinfo{person}{Xingyuan Dai}, \bibinfo{person}{Chen Zhao}, \bibinfo{person}{Xiaoshuang Li}, \bibinfo{person}{Xiao Wang}, {and} \bibinfo{person}{Fei-Yue Wang}.} \bibinfo{year}{2021}\natexlab{}.
\newblock \showarticletitle{Traffic Signal Control Using Offline Reinforcement Learning}. In \bibinfo{booktitle}{\emph{2021 China Automation Congress (CAC)}}. \bibinfo{pages}{8090--8095}.
\newblock
\urldef\tempurl%
\url{https://doi.org/10.1109/CAC53003.2021.9728551}
\showDOI{\tempurl}


\bibitem[Dorfman et~al\mbox{.}(2021)]%
        {OfflineMeta3}
\bibfield{author}{\bibinfo{person}{Ron Dorfman}, \bibinfo{person}{Idan Shenfeld}, {and} \bibinfo{person}{Aviv Tamar}.} \bibinfo{year}{2021}\natexlab{}.
\newblock \showarticletitle{Offline Meta Reinforcement Learning -- Identifiability Challenges and Effective Data Collection Strategies}. In \bibinfo{booktitle}{\emph{Advances in Neural Information Processing Systems}}, \bibfield{editor}{\bibinfo{person}{M.~Ranzato}, \bibinfo{person}{A.~Beygelzimer}, \bibinfo{person}{Y.~Dauphin}, \bibinfo{person}{P.S. Liang}, {and} \bibinfo{person}{J.~Wortman Vaughan}} (Eds.), Vol.~\bibinfo{volume}{34}. \bibinfo{publisher}{Curran Associates, Inc.}, \bibinfo{pages}{4607--4618}.
\newblock
\urldef\tempurl%
\url{https://proceedings.neurips.cc/paper_files/paper/2021/file/248024541dbda1d3fd75fe49d1a4df4d-Paper.pdf}
\showURL{%
\tempurl}


\bibitem[Ernst et~al\mbox{.}(2005)]%
        {batchRL}
\bibfield{author}{\bibinfo{person}{Damien Ernst}, \bibinfo{person}{Pierre Geurts}, {and} \bibinfo{person}{Louis Wehenkel}.} \bibinfo{year}{2005}\natexlab{}.
\newblock \showarticletitle{Tree-Based Batch Mode Reinforcement Learning}.
\newblock \bibinfo{journal}{\emph{J. Mach. Learn. Res.}}  \bibinfo{volume}{6} (\bibinfo{date}{dec} \bibinfo{year}{2005}), \bibinfo{pages}{503–556}.
\newblock
\showISSN{1532-4435}


\bibitem[Eysenbach et~al\mbox{.}(2020a)]%
        {Eysenbach2020}
\bibfield{author}{\bibinfo{person}{Benjamin Eysenbach}, \bibinfo{person}{Xinyang Geng}, \bibinfo{person}{Sergey Levine}, {and} \bibinfo{person}{Ruslan Salakhutdinov}.} \bibinfo{year}{2020}\natexlab{a}.
\newblock \showarticletitle{Rewriting History with Inverse {RL:} Hindsight Inference for Policy Improvement}.
\newblock \bibinfo{journal}{\emph{CoRR}}  \bibinfo{volume}{abs/2002.11089} (\bibinfo{year}{2020}).
\newblock
\showeprint[arXiv]{2002.11089}
\urldef\tempurl%
\url{https://arxiv.org/abs/2002.11089}
\showURL{%
\tempurl}


\bibitem[Eysenbach et~al\mbox{.}(2020b)]%
        {MultiOffline2}
\bibfield{author}{\bibinfo{person}{Benjamin Eysenbach}, \bibinfo{person}{Xinyang Geng}, \bibinfo{person}{Sergey Levine}, {and} \bibinfo{person}{Ruslan Salakhutdinov}.} \bibinfo{year}{2020}\natexlab{b}.
\newblock \showarticletitle{Rewriting History with Inverse {RL:} Hindsight Inference for Policy Improvement}.
\newblock \bibinfo{journal}{\emph{CoRR}}  \bibinfo{volume}{abs/2002.11089} (\bibinfo{year}{2020}).
\newblock
\showeprint[arXiv]{2002.11089}
\urldef\tempurl%
\url{https://arxiv.org/abs/2002.11089}
\showURL{%
\tempurl}


\bibitem[Finn et~al\mbox{.}(2017)]%
        {MAML}
\bibfield{author}{\bibinfo{person}{Chelsea Finn}, \bibinfo{person}{Pieter Abbeel}, {and} \bibinfo{person}{Sergey Levine}.} \bibinfo{year}{2017}\natexlab{}.
\newblock \showarticletitle{Model-Agnostic Meta-Learning for Fast Adaptation of Deep Networks}. In \bibinfo{booktitle}{\emph{Proceedings of the 34th International Conference on Machine Learning - Volume 70}}. \bibinfo{pages}{1126–1135}.
\newblock


\bibitem[Fujimoto et~al\mbox{.}(2018)]%
        {Fujimoto2018}
\bibfield{author}{\bibinfo{person}{Scott Fujimoto}, \bibinfo{person}{David Meger}, {and} \bibinfo{person}{Doina Precup}.} \bibinfo{year}{2018}\natexlab{}.
\newblock \showarticletitle{Off-Policy Deep Reinforcement Learning without Exploration}.
\newblock \bibinfo{journal}{\emph{CoRR}}  \bibinfo{volume}{abs/1812.02900} (\bibinfo{year}{2018}).
\newblock
\showeprint[arXiv]{1812.02900}
\urldef\tempurl%
\url{http://arxiv.org/abs/1812.02900}
\showURL{%
\tempurl}


\bibitem[Goodfellow et~al\mbox{.}(2014)]%
        {GAN}
\bibfield{author}{\bibinfo{person}{Ian Goodfellow}, \bibinfo{person}{Jean Pouget-Abadie}, \bibinfo{person}{Mehdi Mirza}, \bibinfo{person}{Bing Xu}, \bibinfo{person}{David Warde-Farley}, \bibinfo{person}{Sherjil Ozair}, \bibinfo{person}{Aaron Courville}, {and} \bibinfo{person}{Yoshua Bengio}.} \bibinfo{year}{2014}\natexlab{}.
\newblock \showarticletitle{Generative Adversarial Nets}.
\newblock In \bibinfo{booktitle}{\emph{NeurIPS}}.
\newblock


\bibitem[Haarnoja et~al\mbox{.}(2018)]%
        {SAC}
\bibfield{author}{\bibinfo{person}{Tuomas Haarnoja}, \bibinfo{person}{Aurick Zhou}, \bibinfo{person}{Pieter Abbeel}, {and} \bibinfo{person}{Sergey Levine}.} \bibinfo{year}{2018}\natexlab{}.
\newblock \bibinfo{title}{Soft Actor-Critic: Off-Policy Maximum Entropy Deep Reinforcement Learning with a Stochastic Actor}.
\newblock
\newblock
\showeprint[arxiv]{1801.01290}~[cs.LG]


\bibitem[Hessel et~al\mbox{.}(2019)]%
        {hessel2019multi}
\bibfield{author}{\bibinfo{person}{Matteo Hessel}, \bibinfo{person}{Joseph Modayil}, \bibinfo{person}{Hado~van Hasselt}, \bibinfo{person}{Tom Schaul}, \bibinfo{person}{Georg Ostrovski}, \bibinfo{person}{Will Dabney}, \bibinfo{person}{Dan Horgan}, \bibinfo{person}{Bilal Piot}, \bibinfo{person}{Mohammad Azar}, {and} \bibinfo{person}{David Silver}.} \bibinfo{year}{2019}\natexlab{}.
\newblock \showarticletitle{Multi-task deep reinforcement learning with popart}.
\newblock \bibinfo{journal}{\emph{arXiv preprint arXiv:1901.04465}} (\bibinfo{year}{2019}).
\newblock


\bibitem[Ho and Ermon(2016)]%
        {gail}
\bibfield{author}{\bibinfo{person}{Jonathan Ho} {and} \bibinfo{person}{Stefano Ermon}.} \bibinfo{year}{2016}\natexlab{}.
\newblock \showarticletitle{Generative adversarial imitation learning}.
\newblock \bibinfo{journal}{\emph{Advances in neural information processing systems}}  \bibinfo{volume}{29} (\bibinfo{year}{2016}).
\newblock


\bibitem[Jaques et~al\mbox{.}(2019)]%
        {Jaques2019}
\bibfield{author}{\bibinfo{person}{Natasha Jaques}, \bibinfo{person}{Asma Ghandeharioun}, \bibinfo{person}{Judy~Hanwen Shen}, \bibinfo{person}{Craig Ferguson}, \bibinfo{person}{{\`{A}}gata Lapedriza}, \bibinfo{person}{Noah Jones}, \bibinfo{person}{Shixiang Gu}, {and} \bibinfo{person}{Rosalind~W. Picard}.} \bibinfo{year}{2019}\natexlab{}.
\newblock \showarticletitle{Way Off-Policy Batch Deep Reinforcement Learning of Implicit Human Preferences in Dialog}.
\newblock \bibinfo{journal}{\emph{CoRR}}  \bibinfo{volume}{abs/1907.00456} (\bibinfo{year}{2019}).
\newblock
\showeprint[arXiv]{1907.00456}
\urldef\tempurl%
\url{http://arxiv.org/abs/1907.00456}
\showURL{%
\tempurl}


\bibitem[Kalashnikov et~al\mbox{.}(2018)]%
        {Kalashnikov2018}
\bibfield{author}{\bibinfo{person}{Dmitry Kalashnikov}, \bibinfo{person}{Alex Irpan}, \bibinfo{person}{Peter Pastor}, \bibinfo{person}{Julian Ibarz}, \bibinfo{person}{Alexander Herzog}, \bibinfo{person}{Eric Jang}, \bibinfo{person}{Deirdre Quillen}, \bibinfo{person}{Ethan Holly}, \bibinfo{person}{Mrinal Kalakrishnan}, \bibinfo{person}{Vincent Vanhoucke}, {and} \bibinfo{person}{Sergey Levine}.} \bibinfo{year}{2018}\natexlab{}.
\newblock \showarticletitle{QT-Opt: Scalable Deep Reinforcement Learning for Vision-Based Robotic Manipulation}.
\newblock \bibinfo{journal}{\emph{CoRR}}  \bibinfo{volume}{abs/1806.10293} (\bibinfo{year}{2018}).
\newblock
\showeprint[arXiv]{1806.10293}
\urldef\tempurl%
\url{http://arxiv.org/abs/1806.10293}
\showURL{%
\tempurl}


\bibitem[Kalashnikov et~al\mbox{.}(2021)]%
        {MT-Opt}
\bibfield{author}{\bibinfo{person}{Dmitry Kalashnikov}, \bibinfo{person}{Jacob Varley}, \bibinfo{person}{Yevgen Chebotar}, \bibinfo{person}{Benjamin Swanson}, \bibinfo{person}{Rico Jonschkowski}, \bibinfo{person}{Chelsea Finn}, \bibinfo{person}{Sergey Levine}, {and} \bibinfo{person}{Karol Hausman}.} \bibinfo{year}{2021}\natexlab{}.
\newblock \showarticletitle{MT-Opt: Continuous Multi-Task Robotic Reinforcement Learning at Scale}.
\newblock \bibinfo{journal}{\emph{CoRR}}  \bibinfo{volume}{abs/2104.08212} (\bibinfo{year}{2021}).
\newblock
\showeprint[arXiv]{2104.08212}
\urldef\tempurl%
\url{https://arxiv.org/abs/2104.08212}
\showURL{%
\tempurl}


\bibitem[Kidambi et~al\mbox{.}(2020)]%
        {MOReL}
\bibfield{author}{\bibinfo{person}{Rahul Kidambi}, \bibinfo{person}{Aravind Rajeswaran}, \bibinfo{person}{Praneeth Netrapalli}, {and} \bibinfo{person}{Thorsten Joachims}.} \bibinfo{year}{2020}\natexlab{}.
\newblock \showarticletitle{MOReL: Model-Based Offline Reinforcement Learning}. In \bibinfo{booktitle}{\emph{Proceedings of the 34th International Conference on Neural Information Processing Systems}} (Vancouver, BC, Canada) \emph{(\bibinfo{series}{NIPS'20})}. \bibinfo{publisher}{Curran Associates Inc.}, \bibinfo{address}{Red Hook, NY, USA}, Article \bibinfo{articleno}{1830}, \bibinfo{numpages}{14}~pages.
\newblock
\showISBNx{9781713829546}


\bibitem[Kostrikov et~al\mbox{.}(2021)]%
        {Kostrikov2021}
\bibfield{author}{\bibinfo{person}{Ilya Kostrikov}, \bibinfo{person}{Jonathan Tompson}, \bibinfo{person}{Rob Fergus}, {and} \bibinfo{person}{Ofir Nachum}.} \bibinfo{year}{2021}\natexlab{}.
\newblock \showarticletitle{Offline Reinforcement Learning with Fisher Divergence Critic Regularization}.
\newblock \bibinfo{journal}{\emph{CoRR}}  \bibinfo{volume}{abs/2103.08050} (\bibinfo{year}{2021}).
\newblock
\showeprint[arXiv]{2103.08050}
\urldef\tempurl%
\url{https://arxiv.org/abs/2103.08050}
\showURL{%
\tempurl}


\bibitem[Kumar et~al\mbox{.}(2019)]%
        {Kumar2019}
\bibfield{author}{\bibinfo{person}{Aviral Kumar}, \bibinfo{person}{Justin Fu}, \bibinfo{person}{George Tucker}, {and} \bibinfo{person}{Sergey Levine}.} \bibinfo{year}{2019}\natexlab{}.
\newblock \showarticletitle{Stabilizing Off-Policy Q-Learning via Bootstrapping Error Reduction}.
\newblock \bibinfo{journal}{\emph{CoRR}}  \bibinfo{volume}{abs/1906.00949} (\bibinfo{year}{2019}).
\newblock
\showeprint[arXiv]{1906.00949}
\urldef\tempurl%
\url{http://arxiv.org/abs/1906.00949}
\showURL{%
\tempurl}


\bibitem[Kumar et~al\mbox{.}(2020)]%
        {CQL}
\bibfield{author}{\bibinfo{person}{Aviral Kumar}, \bibinfo{person}{Aurick Zhou}, \bibinfo{person}{George Tucker}, {and} \bibinfo{person}{Sergey Levine}.} \bibinfo{year}{2020}\natexlab{}.
\newblock \showarticletitle{Conservative Q-Learning for Offline Reinforcement Learning}. In \bibinfo{booktitle}{\emph{Proceedings of the 34th International Conference on Neural Information Processing Systems}} (Vancouver, BC, Canada) \emph{(\bibinfo{series}{NIPS'20})}. \bibinfo{publisher}{Curran Associates Inc.}, \bibinfo{address}{Red Hook, NY, USA}, Article \bibinfo{articleno}{100}, \bibinfo{numpages}{13}~pages.
\newblock
\showISBNx{9781713829546}


\bibitem[Kunjir and Chawla(2022)]%
        {Kunjir2022}
\bibfield{author}{\bibinfo{person}{Mayuresh Kunjir} {and} \bibinfo{person}{Sanjay Chawla}.} \bibinfo{year}{2022}\natexlab{}.
\newblock \showarticletitle{Offline Reinforcement Learning for Road Traffic Control}.
\newblock \bibinfo{journal}{\emph{CoRR}}  \bibinfo{volume}{abs/2201.02381} (\bibinfo{year}{2022}).
\newblock
\showeprint[arXiv]{2201.02381}
\urldef\tempurl%
\url{https://arxiv.org/abs/2201.02381}
\showURL{%
\tempurl}


\bibitem[Lange et~al\mbox{.}(2012)]%
        {batchRL2}
\bibfield{author}{\bibinfo{person}{Sascha Lange}, \bibinfo{person}{Thomas Gabel}, {and} \bibinfo{person}{Martin Riedmiller}.} \bibinfo{year}{2012}\natexlab{}.
\newblock \bibinfo{booktitle}{\emph{Batch Reinforcement Learning}}.
\newblock \bibinfo{publisher}{Springer Berlin Heidelberg}, \bibinfo{address}{Berlin, Heidelberg}, \bibinfo{pages}{45--73}.
\newblock
\showISBNx{978-3-642-27645-3}
\urldef\tempurl%
\url{https://doi.org/10.1007/978-3-642-27645-3_2}
\showDOI{\tempurl}


\bibitem[Levine et~al\mbox{.}(2020)]%
        {OfflineTutorial}
\bibfield{author}{\bibinfo{person}{Sergey Levine}, \bibinfo{person}{Aviral Kumar}, \bibinfo{person}{George Tucker}, {and} \bibinfo{person}{Justin Fu}.} \bibinfo{year}{2020}\natexlab{}.
\newblock \showarticletitle{Offline Reinforcement Learning: Tutorial, Review, and Perspectives on Open Problems}.
\newblock \bibinfo{journal}{\emph{CoRR}}  \bibinfo{volume}{abs/2005.01643} (\bibinfo{year}{2020}).
\newblock
\showeprint[arXiv]{2005.01643}
\urldef\tempurl%
\url{https://arxiv.org/abs/2005.01643}
\showURL{%
\tempurl}


\bibitem[Li et~al\mbox{.}(2023)]%
        {Li2023}
\bibfield{author}{\bibinfo{person}{Jianxiong Li}, \bibinfo{person}{Shichao Lin}, \bibinfo{person}{Tianyu Shi}, \bibinfo{person}{Chujie Tian}, \bibinfo{person}{Yu Mei}, \bibinfo{person}{Jian Song}, \bibinfo{person}{Xianyuan Zhan}, {and} \bibinfo{person}{Ruimin Li}.} \bibinfo{year}{2023}\natexlab{}.
\newblock \bibinfo{title}{A Fully Data-Driven Approach for Realistic Traffic Signal Control Using Offline Reinforcement Learning}.
\newblock
\newblock
\showeprint[arxiv]{2311.15920}~[cs.AI]


\bibitem[Lin et~al\mbox{.}(2019)]%
        {Lin2019}
\bibfield{author}{\bibinfo{person}{Xingyu Lin}, \bibinfo{person}{Harjatin~Singh Baweja}, {and} \bibinfo{person}{David Held}.} \bibinfo{year}{2019}\natexlab{}.
\newblock \showarticletitle{Reinforcement Learning without Ground-Truth State}.
\newblock \bibinfo{journal}{\emph{CoRR}}  \bibinfo{volume}{abs/1905.07866} (\bibinfo{year}{2019}).
\newblock
\showeprint[arXiv]{1905.07866}
\urldef\tempurl%
\url{http://arxiv.org/abs/1905.07866}
\showURL{%
\tempurl}


\bibitem[Liu et~al\mbox{.}(2019)]%
        {Liu2019}
\bibfield{author}{\bibinfo{person}{Hao Liu}, \bibinfo{person}{Alexander Trott}, \bibinfo{person}{Richard Socher}, {and} \bibinfo{person}{Caiming Xiong}.} \bibinfo{year}{2019}\natexlab{}.
\newblock \showarticletitle{Competitive Experience Replay}.
\newblock \bibinfo{journal}{\emph{CoRR}}  \bibinfo{volume}{abs/1902.00528} (\bibinfo{year}{2019}).
\newblock
\showeprint[arXiv]{1902.00528}
\urldef\tempurl%
\url{http://arxiv.org/abs/1902.00528}
\showURL{%
\tempurl}


\bibitem[Mitchell et~al\mbox{.}(2020)]%
        {OfflineMeta1}
\bibfield{author}{\bibinfo{person}{Eric Mitchell}, \bibinfo{person}{Rafael Rafailov}, \bibinfo{person}{Xue~Bin Peng}, \bibinfo{person}{Sergey Levine}, {and} \bibinfo{person}{Chelsea Finn}.} \bibinfo{year}{2020}\natexlab{}.
\newblock \showarticletitle{Offline Meta-Reinforcement Learning with Advantage Weighting}.
\newblock \bibinfo{journal}{\emph{CoRR}}  \bibinfo{volume}{abs/2008.06043} (\bibinfo{year}{2020}).
\newblock
\showeprint[arXiv]{2008.06043}
\urldef\tempurl%
\url{https://arxiv.org/abs/2008.06043}
\showURL{%
\tempurl}


\bibitem[Pan et~al\mbox{.}(2020)]%
        {xgail}
\bibfield{author}{\bibinfo{person}{Menghai Pan}, \bibinfo{person}{Weixiao Huang}, \bibinfo{person}{Yanhua Li}, \bibinfo{person}{Xun Zhou}, {and} \bibinfo{person}{Jun Luo}.} \bibinfo{year}{2020}\natexlab{}.
\newblock \showarticletitle{XGAIL: Explainable Generative Adversarial Imitation Learning for Explainable Human Decision Analysis}. In \bibinfo{booktitle}{\emph{KDD, 2020}}.
\newblock


\bibitem[Peng et~al\mbox{.}(2019)]%
        {Peng2019}
\bibfield{author}{\bibinfo{person}{Xue~Bin Peng}, \bibinfo{person}{Aviral Kumar}, \bibinfo{person}{Grace Zhang}, {and} \bibinfo{person}{Sergey Levine}.} \bibinfo{year}{2019}\natexlab{}.
\newblock \showarticletitle{Advantage-Weighted Regression: Simple and Scalable Off-Policy Reinforcement Learning}.
\newblock \bibinfo{journal}{\emph{CoRR}}  \bibinfo{volume}{abs/1910.00177} (\bibinfo{year}{2019}).
\newblock
\showeprint[arXiv]{1910.00177}
\urldef\tempurl%
\url{http://arxiv.org/abs/1910.00177}
\showURL{%
\tempurl}


\bibitem[Pong et~al\mbox{.}(2021)]%
        {OfflineMeta2}
\bibfield{author}{\bibinfo{person}{Vitchyr~H. Pong}, \bibinfo{person}{Ashvin Nair}, \bibinfo{person}{Laura~M. Smith}, \bibinfo{person}{Catherine Huang}, {and} \bibinfo{person}{Sergey Levine}.} \bibinfo{year}{2021}\natexlab{}.
\newblock \showarticletitle{Offline Meta-Reinforcement Learning with Online Self-Supervision}.
\newblock \bibinfo{journal}{\emph{CoRR}}  \bibinfo{volume}{abs/2107.03974} (\bibinfo{year}{2021}).
\newblock
\showeprint[arXiv]{2107.03974}
\urldef\tempurl%
\url{https://arxiv.org/abs/2107.03974}
\showURL{%
\tempurl}


\bibitem[Rafailov et~al\mbox{.}(2020)]%
        {Rafailov2020}
\bibfield{author}{\bibinfo{person}{Rafael Rafailov}, \bibinfo{person}{Tianhe Yu}, \bibinfo{person}{Aravind Rajeswaran}, {and} \bibinfo{person}{Chelsea Finn}.} \bibinfo{year}{2020}\natexlab{}.
\newblock \showarticletitle{Offline Reinforcement Learning from Images with Latent Space Models}.
\newblock \bibinfo{journal}{\emph{CoRR}}  \bibinfo{volume}{abs/2012.11547} (\bibinfo{year}{2020}).
\newblock
\showeprint[arXiv]{2012.11547}
\urldef\tempurl%
\url{https://arxiv.org/abs/2012.11547}
\showURL{%
\tempurl}


\bibitem[Schroff et~al\mbox{.}(2015)]%
        {triplet_loss}
\bibfield{author}{\bibinfo{person}{Florian Schroff}, \bibinfo{person}{Dmitry Kalenichenko}, {and} \bibinfo{person}{James Philbin}.} \bibinfo{year}{2015}\natexlab{}.
\newblock \showarticletitle{FaceNet: A unified embedding for face recognition and clustering}. In \bibinfo{booktitle}{\emph{2015 IEEE Conference on Computer Vision and Pattern Recognition (CVPR)}}. \bibinfo{publisher}{IEEE}.
\newblock
\urldef\tempurl%
\url{https://doi.org/10.1109/cvpr.2015.7298682}
\showDOI{\tempurl}


\bibitem[Shah et~al\mbox{.}(2022)]%
        {shah2022offline}
\bibfield{author}{\bibinfo{person}{Dhruv Shah}, \bibinfo{person}{Arjun Bhorkar}, \bibinfo{person}{Hrishit Leen}, \bibinfo{person}{Ilya Kostrikov}, \bibinfo{person}{Nicholas Rhinehart}, {and} \bibinfo{person}{Sergey Levine}.} \bibinfo{year}{2022}\natexlab{}.
\newblock \showarticletitle{Offline Reinforcement Learning for Visual Navigation}. In \bibinfo{booktitle}{\emph{6th Annual Conference on Robot Learning}}.
\newblock
\urldef\tempurl%
\url{https://openreview.net/forum?id=uhIfIEIiWm_}
\showURL{%
\tempurl}


\bibitem[Shi et~al\mbox{.}(2021)]%
        {Shi2021}
\bibfield{author}{\bibinfo{person}{Tianyu Shi}, \bibinfo{person}{Dong Chen}, \bibinfo{person}{Kaian Chen}, {and} \bibinfo{person}{Zhaojian Li}.} \bibinfo{year}{2021}\natexlab{}.
\newblock \showarticletitle{Offline Reinforcement Learning for Autonomous Driving with Safety and Exploration Enhancement}.
\newblock \bibinfo{journal}{\emph{CoRR}}  \bibinfo{volume}{abs/2110.07067} (\bibinfo{year}{2021}).
\newblock
\showeprint[arXiv]{2110.07067}
\urldef\tempurl%
\url{https://arxiv.org/abs/2110.07067}
\showURL{%
\tempurl}


\bibitem[Siegel et~al\mbox{.}(2020)]%
        {Siegel2020}
\bibfield{author}{\bibinfo{person}{Noah~Y. Siegel}, \bibinfo{person}{Jost~Tobias Springenberg}, \bibinfo{person}{Felix Berkenkamp}, \bibinfo{person}{Abbas Abdolmaleki}, \bibinfo{person}{Michael Neunert}, \bibinfo{person}{Thomas Lampe}, \bibinfo{person}{Roland Hafner}, \bibinfo{person}{Nicolas Heess}, {and} \bibinfo{person}{Martin~A. Riedmiller}.} \bibinfo{year}{2020}\natexlab{}.
\newblock \showarticletitle{Keep Doing What Worked: Behavioral Modelling Priors for Offline Reinforcement Learning}.
\newblock \bibinfo{journal}{\emph{CoRR}}  \bibinfo{volume}{abs/2002.08396} (\bibinfo{year}{2020}).
\newblock
\showeprint[arXiv]{2002.08396}
\urldef\tempurl%
\url{https://arxiv.org/abs/2002.08396}
\showURL{%
\tempurl}


\bibitem[Sun et~al\mbox{.}(2019)]%
        {Sun2019}
\bibfield{author}{\bibinfo{person}{Hao Sun}, \bibinfo{person}{Zhizhong Li}, \bibinfo{person}{Xiaotong Liu}, \bibinfo{person}{Dahua Lin}, {and} \bibinfo{person}{Bolei Zhou}.} \bibinfo{year}{2019}\natexlab{}.
\newblock \showarticletitle{Policy Continuation with Hindsight Inverse Dynamics}.
\newblock \bibinfo{journal}{\emph{CoRR}}  \bibinfo{volume}{abs/1910.14055} (\bibinfo{year}{2019}).
\newblock
\showeprint[arXiv]{1910.14055}
\urldef\tempurl%
\url{http://arxiv.org/abs/1910.14055}
\showURL{%
\tempurl}


\bibitem[Tessler et~al\mbox{.}(2017)]%
        {tessler2017deep}
\bibfield{author}{\bibinfo{person}{Chen Tessler}, \bibinfo{person}{Shahar Givony}, \bibinfo{person}{Tom Zahavy}, \bibinfo{person}{Daniel~J Mankowitz}, {and} \bibinfo{person}{Shie Mannor}.} \bibinfo{year}{2017}\natexlab{}.
\newblock \showarticletitle{Deep multi-task learning with low level tasks supervised at lower layers}.
\newblock \bibinfo{journal}{\emph{arXiv preprint arXiv:1704.05098}} (\bibinfo{year}{2017}).
\newblock


\bibitem[Vaswani et~al\mbox{.}(2017)]%
        {transformer}
\bibfield{author}{\bibinfo{person}{Ashish Vaswani}, \bibinfo{person}{Noam Shazeer}, \bibinfo{person}{Niki Parmar}, \bibinfo{person}{Jakob Uszkoreit}, \bibinfo{person}{Llion Jones}, \bibinfo{person}{Aidan~N. Gomez}, \bibinfo{person}{Lukasz Kaiser}, {and} \bibinfo{person}{Illia Polosukhin}.} \bibinfo{year}{2017}\natexlab{}.
\newblock \showarticletitle{Attention Is All You Need}.
\newblock \bibinfo{journal}{\emph{CoRR}} (\bibinfo{year}{2017}).
\newblock


\bibitem[Venkatraman et~al\mbox{.}(2015)]%
        {Venkatraman2015}
\bibfield{author}{\bibinfo{person}{Arun Venkatraman}, \bibinfo{person}{Martial Hebert}, {and} \bibinfo{person}{J.. Bagnell}.} \bibinfo{year}{2015}\natexlab{}.
\newblock \showarticletitle{Improving Multi-Step Prediction of Learned Time Series Models}.
\newblock \bibinfo{journal}{\emph{Proceedings of the AAAI Conference on Artificial Intelligence}} \bibinfo{volume}{29}, \bibinfo{number}{1} (\bibinfo{date}{Feb.} \bibinfo{year}{2015}).
\newblock
\urldef\tempurl%
\url{https://doi.org/10.1609/aaai.v29i1.9590}
\showDOI{\tempurl}


\bibitem[Wang et~al\mbox{.}(2018)]%
        {Wang2018}
\bibfield{author}{\bibinfo{person}{Lu Wang}, \bibinfo{person}{Wei Zhang}, \bibinfo{person}{Xiaofeng He}, {and} \bibinfo{person}{Hongyuan Zha}.} \bibinfo{year}{2018}\natexlab{}.
\newblock \showarticletitle{Supervised Reinforcement Learning with Recurrent Neural Network for Dynamic Treatment Recommendation}.
\newblock \bibinfo{journal}{\emph{CoRR}}  \bibinfo{volume}{abs/1807.01473} (\bibinfo{year}{2018}).
\newblock
\showeprint[arXiv]{1807.01473}
\urldef\tempurl%
\url{http://arxiv.org/abs/1807.01473}
\showURL{%
\tempurl}


\bibitem[Wu et~al\mbox{.}(2019)]%
        {Wu2019}
\bibfield{author}{\bibinfo{person}{Yifan Wu}, \bibinfo{person}{George Tucker}, {and} \bibinfo{person}{Ofir Nachum}.} \bibinfo{year}{2019}\natexlab{}.
\newblock \showarticletitle{Behavior Regularized Offline Reinforcement Learning}.
\newblock \bibinfo{journal}{\emph{CoRR}}  \bibinfo{volume}{abs/1911.11361} (\bibinfo{year}{2019}).
\newblock
\showeprint[arXiv]{1911.11361}
\urldef\tempurl%
\url{http://arxiv.org/abs/1911.11361}
\showURL{%
\tempurl}


\bibitem[Yu et~al\mbox{.}(2022)]%
        {UDS}
\bibfield{author}{\bibinfo{person}{Tianhe Yu}, \bibinfo{person}{Aviral Kumar}, \bibinfo{person}{Yevgen Chebotar}, \bibinfo{person}{Karol Hausman}, \bibinfo{person}{Chelsea Finn}, {and} \bibinfo{person}{Sergey Levine}.} \bibinfo{year}{2022}\natexlab{}.
\newblock \showarticletitle{How to Leverage Unlabeled Data in Offline Reinforcement Learning}.
\newblock \bibinfo{journal}{\emph{CoRR}}  \bibinfo{volume}{abs/2202.01741} (\bibinfo{year}{2022}).
\newblock
\showeprint[arXiv]{2202.01741}
\urldef\tempurl%
\url{https://arxiv.org/abs/2202.01741}
\showURL{%
\tempurl}


\bibitem[Yu et~al\mbox{.}(2021)]%
        {CDS}
\bibfield{author}{\bibinfo{person}{Tianhe Yu}, \bibinfo{person}{Aviral Kumar}, \bibinfo{person}{Yevgen Chebotar}, \bibinfo{person}{Karol Hausman}, \bibinfo{person}{Sergey Levine}, {and} \bibinfo{person}{Chelsea Finn}.} \bibinfo{year}{2021}\natexlab{}.
\newblock \showarticletitle{Conservative Data Sharing for Multi-Task Offline Reinforcement Learning}.
\newblock \bibinfo{journal}{\emph{CoRR}}  \bibinfo{volume}{abs/2109.08128} (\bibinfo{year}{2021}).
\newblock
\showeprint[arXiv]{2109.08128}
\urldef\tempurl%
\url{https://arxiv.org/abs/2109.08128}
\showURL{%
\tempurl}


\bibitem[Yu et~al\mbox{.}(2020)]%
        {MOPO}
\bibfield{author}{\bibinfo{person}{Tianhe Yu}, \bibinfo{person}{Garrett Thomas}, \bibinfo{person}{Lantao Yu}, \bibinfo{person}{Stefano Ermon}, \bibinfo{person}{James~Y Zou}, \bibinfo{person}{Sergey Levine}, \bibinfo{person}{Chelsea Finn}, {and} \bibinfo{person}{Tengyu Ma}.} \bibinfo{year}{2020}\natexlab{}.
\newblock \showarticletitle{MOPO: Model-based Offline Policy Optimization}. In \bibinfo{booktitle}{\emph{Advances in Neural Information Processing Systems}}, Vol.~\bibinfo{volume}{33}. \bibinfo{publisher}{Curran Associates, Inc.}, \bibinfo{pages}{14129--14142}.
\newblock


\bibitem[Zhang et~al\mbox{.}(2020a)]%
        {cgail}
\bibfield{author}{\bibinfo{person}{Xin Zhang}, \bibinfo{person}{Yanhua Li}, \bibinfo{person}{Xun Zhou}, {and} \bibinfo{person}{Jun Luo}.} \bibinfo{year}{2020}\natexlab{a}.
\newblock \showarticletitle{cGAIL: Conditional Generative Adversarial Imitation Learning—An Application in Taxi Drivers’ Strategy Learning}.
\newblock \bibinfo{journal}{\emph{IEEE TBD}} (\bibinfo{year}{2020}).
\newblock


\bibitem[Zhang et~al\mbox{.}(2020b)]%
        {traj_gail}
\bibfield{author}{\bibinfo{person}{Xin Zhang}, \bibinfo{person}{Yanhua Li}, \bibinfo{person}{Xun Zhou}, \bibinfo{person}{Ziming Zhang}, {and} \bibinfo{person}{Jun Luo}.} \bibinfo{year}{2020}\natexlab{b}.
\newblock \showarticletitle{TrajGAIL: Trajectory Generative Adversarial Imitation Learning for Long-Term Decision Analysis}. In \bibinfo{booktitle}{\emph{ICDM'20}}.
\newblock


\bibitem[Zhang et~al\mbox{.}(2023)]%
        {stmgail}
\bibfield{author}{\bibinfo{person}{Yingxue Zhang}, \bibinfo{person}{Yanhua Li}, \bibinfo{person}{Xun Zhou}, \bibinfo{person}{Ziming Zhang}, {and} \bibinfo{person}{Jun Luo}.} \bibinfo{year}{2023}\natexlab{}.
\newblock \bibinfo{booktitle}{\emph{STM-GAIL: Spatial-Temporal Meta-GAIL for Learning Diverse Human Driving Strategies}}.
\newblock


\end{thebibliography}


\appendix

\section{Appendix}
To support the reproducibility, our code and data are released online\footnote{https://github.com/anony11sdf/MODA}.

\subsection{Detailed Settings of Contrastive Data Sharing}

\subsubsection{Contrastive network}
Our dataset contains a large number of taxi drivers seeking passenger trajectories. We train the contrastive network by sampling triplets randomly from the dataset.
Specifically, we first select the target driver, set the size of the sub-trajectory, and the Positive range, and then select the Anchor, and the Positive sub-trajectory in the manner described in the main text. At this point, we iterate through all the trajectories of all the other drivers, and find the the negative sub-trajectory, thus forming the triples. In the training process, for a triple, we forward propagate each of the three sub-trajectories to get embeddings, and then compute the contrastive loss. In the training process, for a triple, we forward propagate each of the three sub-trajectories to get embedding, and then compute the contrastive loss.

It is worth noting that the number of triples composed in this way will be very large, so in order to train the model more efficiently, we will randomly filter the composition of triples to decide whether to use it to train the model or not. Usually, we filter 10\% to 20\%, if the sub-trajectory contains more transitions, it means that we need to filter fewer triples because the GPU memory is limited.

\subsubsection{Sharing data}

Despite we have Contrastive models, sharing data can be done in many ways. We choose Eq.~\eqref{eq:sharing} for data-sharing to improve efficiency.
According to the Contrastive loss, the most intuitive way to share data is to collect data as in the training process (without random filtering), and then form triples. then calculate the Contrastive loss, and then set a threshold to decide whether to share the current negative sub-trajectory that gets this Contrastive loss or not. 
However, in the process of implementation, we found that this way of sharing data consumes a lot of arithmetic power, and at the same time, due to the huge number of triples, it is almost impossible to check all negatives, so it is inefficient. That's why we use the Eq.~\eqref{eq:sharing} description for data analysis. This approach improves the efficiency of the whole data sharing by hundreds of times, and at the same time, it also reduces a hyperparameter (Contrastive loss sharing threshold). Also, In the end, the data-sharing approach we adopted works well.

\subsection{Detailed Settings of Model-based Multi-task offline RL with Robust MDP}
We use transitions from effective dataset $\mathcal{D}_{i}^{\mathrm{eff}}$ to train GAN models and dynamics models.

\subsubsection{GAN}
In GAN, both the Generator and Discriminator utilize fully connected neural networks with ReLU activation functions. It is configured with a batch size of 32 and trained over 2000 epochs. The learning rates for the generator and discriminator are set to 0.0005 and 0.0002, respectively.

\subsubsection{Dynamics model}
Dynamics model $\hat{P}_\psi$ encompasses two models tasked with predicting rewards and subsequent states, respectively. The learning rate is set to 0.02, with a batch size of 32, and the model is trained for 2000 epochs. 

For state Sub-Model: This model predicts the next state given the current state and action. It has two fully connected layers with ReLU activation functions. The input size is the sum of the state size and action size, and the output size is the state size.

For reward Sub-Model: This model predicts the reward given the current state and action. It has two fully connected layers with ReLU activation functions. The input size is the sum of the state size and action size, and the output size is the reward size. 

\subsubsection{Robust MDP construction}
After completing the training of the GAN and dynamics models, we need to build a Robust MDP by combining the GAN's discriminators with the dynamics models. It works like this: first, we randomize an initial state that is randomly drawn from the entire dataset in a transition Then the policy obtained by the subsequent algorithm generates an action based on this state, and this state and action are fed to the dynamics models, which predicts the forward and next states and forms a transition. this transition is then this transition will be input to the discriminator, which will output 0 or 1 to determine whether the transition is real or not. If the discriminator outputs 0, it means that the current transition is false, and the environment will be automatically set to the halt state and return to done, which represents the end of the current episode. This halt state will set the reward to a very small value as a penalty.

About why the initial state is randomly selected instead of randomly generated. Because the environment we simulate is a continuous environment with very high dimensionality, the randomly generated state has a high probability of being unrealistic and also has a high probability of being different from the distribution of the data that the model has learned, which will cause the Robust MDP to end the current episodes immediately.

\subsubsection{Soft-Actor-Critic (SAC)} This algorithm contains 2 models, SACActor and SACCritic. The SACActor model is responsible for predicting action logits given the state, while the SACCritic model estimates Q-values given the state-action pair. Additionally, a ReplayBuffer class is implemented to store and sample experiences for training. The hyperparameters used in training include the total number of episodes (20000), batch size (64), discount factor (gamma = 0.99), soft update coefficient (tau = 0.005), and entropy regularization coefficient (alpha = 0.2). The replay buffer capacity is set to 300000.

\subsection{Passenger-seeking simulation and data selection details}
Our dataset comprises trajectories collected from 17,877 drivers, however, the data quality of all these drivers varies, many of them suffer from incomplete trajectories. And in our evaluation, we use many transitions' data from whole dataset for environment simulation, and we use 20 drivers (i.e., 20 tasks) for training all offline RL models (including MODA and all baselines), these drivers provide the most complete and good quality datasets. Briefly speaking, we are solving a 20-task offline RL problem in our experiment instead of a 17,877-task problem.
To simulate the real environment of the city aiming to test the policies learned with our MODA and all baselines, We created a simulated passenger-seeking environment based on the above taxi trajectory data. Specifically this model is an ensemble model, which contains ensemble size dynamics models, each of which has the same structure as previously mentioned. In our experiments, we set the ensemble size to 5. Most importantly, the data we use to train the ensemble model contains all the transitions in the dataset, which is several times more than the amount of data used to train MODA and other baselines, so this model can better simulate the passer- seeking environment.

\subsection{Baselines}

For CQL, BCQ and BEAR, since they are all offline reinforcement learning algorithms, they are trained directly on the dataset and do not need to be deployed in a certain environment like SAC.

\begin{itemize}[nosep, leftmargin=*]

    \item \textit{CQL}. This model consists of three fully connected layers, each followed by a rectified linear unit (ReLU) activation function. The input dimension of the network corresponds to the state space, and the output dimension corresponds to the action space. The network is trained using the Adam optimizer with a learning rate of 0.001. Additionally, the model sets alpha to 0.1, and trains 1000 epochs with batch size 64.

    \item \textit{BCQ} This model includes two neural networks: a Q-network and an action generator. Both networks consist of three fully connected layers with ReLU activation functions. The Q-network takes the state as input and outputs Q-values for each action, while the action generator takes the state as input and outputs a probability distribution over actions using the softmax function. The model is trained using behavior cloning from expert demonstrations, with data loaded from the provided buffer path. The training process runs for 1000 epochs with a learning rate of 0.0001.

    \item \textit{BEAR}. This model consists of two neural networks: a Q-network and a policy network. The Q-network has three fully connected layers with ReLU activation functions, taking both state and action as input. It outputs Q-values estimating the expected cumulative future rewards for each action. The policy network also has three fully connected layers with ReLU activation functions, taking only the state as input and outputting a probability distribution over actions using the softmax function. Both networks are optimized using the Adam optimizer with a learning rate of 0.001. Additionally, the model employs a discount factor of 0.99 for future rewards and sets the MMD loss weight to 0.1. The training loop iterates for 1000 epochs with a batch size of 64.

    \item \textit{MOReL}. We trained ensemble models containing multiple dynamic models in a single driver's data constructed the pessimistic MDP and then deployed the SAC algorithm on that environment.

\end{itemize}

\subsection{Detailed Settings of experiments on Impact of different number of transitions shared by contrastive data sharing}

After getting effective dataset $\mathcal{D}_{i}^{\mathrm{eff}}$ again, the number of transitions it contains is fixed, so for comparison experiments, we labeled the transitions that were shared in, and then randomly chose the specified number of transitions to be used to construct the dataset used as the comparison test. If the specified number is larger than the total number of shared transitions, we select the remaining number of unshared transitions.

\subsection{Discussion on scalability and applicability}

Our framework has good scalability and applicability to various other tasks and applications. 
On a theoretical level, it's a general framework that can be used in a wide range of applications. Besides, as an example, our experiments are an urban application of the real world. Therefore, any real-world urban tasks (e.g., public transportation, bike-sharing, ride-sharing), that have offline datasets,  can leverage our framework. 
From the experimental or application level, based on the dataset of real-world urban tasks, we can abstract many trajectories of many agents active in the same environment based on, for example, GPS records and map information, and then apply our framework on the dataset of these agents.


\end{document}